# Benchmarking Deep Learning Models for Laryngeal Cancer Staging Using the LaryngealCT Dataset

Nivea Roy[1,2], Son N. Tran[1], Atul Sajjanhar[1], K. Devaraja[2*], Prakashini Koteshwara[3], Yong Xiang[1], Divya Rao [4]

[1] School of Information Technology, Deakin University, Burwood, Australia

[2] Department of Head and Neck Surgery, Kasturba Medical College, Manipal Academy of Higher Education, Manipal, India

[3] Department of Radiodiagnosis and Imaging, Kasturba Medical College, Manipal Academy of Higher Education, Manipal, India

[4] Manipal Institute of Technology, Manipal Academy of Higher Education, Manipal, India

[*]**Corresponding author**: devaraja.k@manipal.edu

# Abstract

Laryngeal cancer imaging research lacks standardised public datasets to enable reproducible deep learning (DL) model development. We present LaryngealCT, a curated benchmark of 1,029 computed tomography (CT) scans aggregated from six collections from The Cancer Imaging Archive (TCIA). Uniform 1 mm isotropic volumes of interest encompassing the larynx were extracted using a weakly supervised parameter search framework validated by clinical experts. Six 3D DL architectures (custom 3D CNN, ResNet18/50/101, DenseNet121 and MedicalNet-pretrained ResNet50) were benchmarked on (i) early (Tis–T2) vs. advanced (T3–T4) and (ii) T4 vs. non-T4 classification tasks. On the independent test set, the 3D CNN achieved the strongest overall performance across global and per-class metrics (Accuracy=0.854, F1-macro=0.841) in early vs advanced classification. In the T4 task, AU-ROC values exceeded 0.82 for most models, but sensitivity for T4 disease remained limited (≤0.412), with ResNet101 showing the most promising calibrated T4 recall (0.706). Model explainability assessed using GradCAM++ with thyroid cartilage overlays for the T4 classification task revealed anatomically plausible peri-cartilage activations although spatial overlap remained modest. Through open-source data, pretrained models, and integrated explainability tools, LaryngealCT offers a reproducible foundation for AI-driven research to support future clinical decision-making in laryngeal oncology.

# Introduction

Laryngeal cancer remains one of the most common malignancies of the head and neck, affecting thousands globally each year and contributing significantly to morbidity and mortality[1,2]. Accurate diagnosis and precise tumour staging are pivotal for optimising treatment strategies and improving patient outcomes, yet they remain challenging due to the anatomical and pathological complexity of the larynx and variability in tumour presentation[3,4]. Recent advances in artificial intelligence (AI), and especially deep learning (DL), have transformed medical imaging by enabling automated detection, segmentation, and prognostic modelling across multiple cancer types[5]. Yet the success of these methods depends heavily on access to large, curated, and standardised datasets that enable reproducible model development and robust benchmarking[5–7].

However, in laryngeal oncology, progress is constrained by the absence of dedicated, multi-institutional resources. Contrast-enhanced computed tomography (CE-CT) remains the primary imaging modality for evaluating laryngeal tumours, offering high spatial resolution for assessing cartilage invasion and extralaryngeal spread[3,4,8]. Existing public head-and-neck datasets, including MICCAI HECKTOR and HaN-Seg, address related but distinct problems[7,9]. HECKTOR provides multi-centre Fluorodeoxyglucose (FDG)-Positron Emission Tomography (PET) and Computed Tomography (CT) with expert segmentations of primary tumour and involved nodes in mainly oropharyngeal disease, supporting PET/CT-based lesion segmentation and outcome modelling, but does not supply larynx-focused CT crops or harmonised T-stage labels suitable for dedicated laryngeal cancer staging benchmarks[7] . HaN-Seg, in turn, offers high-quality Computed Tomography/Magnetic Resonance Imaging (CT/MRI) organ-at-risk (OAR) masks for a wide range of head-and-neck structures (including glottic and supraglottic larynx), enabling OAR segmentation research rather than tumour T-staging, and similarly lacks curated laryngeal Volumes of Interest (VOIs) paired with Tumour-Node-Metastasis (TNM) metadata[9]. To our knowledge, LaryngealCT is the first multi-institutional, ROI-focused CT benchmark specifically designed for laryngeal cancer staging rather than generic head-and-neck analysis.

**Table 1 | Summary of AI-based classification studies using CT imaging in laryngeal cancer**

| Study | Sample Size | Input | Model | Task | Performance | Key Limitations |
|---|---|---|---|---|---|---|
| **Santin et al.[10], 2019** | 326 (244 train / 82 test; eval: 189) | 2D CT (JPEG, augmented ×6) | DL (VGG16, transfer learning) | Detect thyroid cartilage abnormality | Test: AUC 0.72 (Sn 0.83, Sp 0.64); Eval: AUC 0.70 | Slice-based; harmonized 2D data; limited generalisability |
| **Guo et al.[11], 2020** | 265 (86 invasion / 179 non-invasion) | CE-CT | Radiomics (LR, SVM-SMOTE) | Predict thyroid cartilage invasion | AUC 0.906 (LR-SVM); AUC 0.876 (LR) | Modest cohort; manual delineation; no external validation |
| **Rao et al.[14], 2023** | 238 (5-fold CV) | CE-CT | Radiomics (LR, SFFS feature selection) | Classify tumour site (supraglottic vs others) | Acc 0.96 | Single-center; no external validation; not 3D VOIs |
| **Chen et al.[18], 2023** | 319 (223 train / 96 test) | CE-CT (venous phase) | Radiomics; DL (ResNet18); Hybrid (Radiomics+DL) | Classify early (T1–T2) vs advanced (T3–T4) | Radiomics: AUC 0.704 (Sn 0.56, Sp 0.83); DL: AUC 0.724; Hybrid: AUC 0.824 (Sn 0.83, Sp 0.85) | Moderate dataset; single-phase CT; no external validation |
| **Hao et al.[13], 2024** | 286 (179 train / 77 test; external val: 30) | Non-contrast CT | DL (ResNet-3Dsml) | Classify thyroid cartilage invasion | Train/Test: AUC 0.844 (Sn 0.56, Sp 0.97, F1 0.67); Val: AUC 0.856 (Sn 0.40, Sp 0.88, F1 0.40) | High specificity but low sensitivity; no data/code release |
| **Takano et al.[16], 2025** | 91 (61 train / 30 test) | CE-CT & Non-CE CT | DL (ResNet101, pretrained) | Classify thyroid cartilage invasion | Without masking: AUC 0.72 (Sn 0.80, Sp 0.70); With masking: AUC 0.82 (Sn 0.80, Sp 0.95) | Small, single-center cohort; limited sensitivity; no per-class metrics |

Sn = Sensitivity; Sp = Specificity; AUC = Area Under the ROC Curve; Acc = Accuracy; DL = Deep Learning; CE-CT = Contrast-Enhanced Computed Tomography.

Previous research has applied radiomics and DL to staging and subsite classification, but these studies are frequently limited by small sample sizes, single-centre cohorts, lack of standardised three-dimensional (3D) regions of interest (ROI), and limited reproducibility due to unavailable data or code[10–19]. Reported performance varies, but external validation and access to landmark datasets remain rare, impeding technical innovation, benchmarking, and clinical translation. Moreover, integration of multimodal and clinical metadata is often incomplete, limiting AI research relevance to real-world workflows. Even within The Cancer Imaging Archive (TCIA) ecosystem, individual collections such as RADCURE, Head-Neck-PET-CT, HEAD-NECK-RADIOMICS-HN1, HNSCC, HNSCC-3DCT-RT and QIN-HEADNECK provide heterogenous whole-neck scans with varying annotation depth and clinical fields, yet none deliver standardised, larynx-centered VOIs with harmonised TNM T-stage labels and a unified preprocessing pipeline explicitly tailored to laryngeal cancer staging CT/MRI[20–26]. The details of major AI-based classification studies using CT imaging in laryngeal cancer are summarised in Table 1.

To address these barriers in the field of laryngeal cancer staging, we present a comprehensive benchmarking of six state-of-the-art 3D deep learning architectures ( a custom 3D CNN, ResNet18/50/101[27], DenseNet121[28], and a MedicalNet-pretrained[29] ResNet50) on two clinically pertinent classification tasks: early (Tis/T1/T2) versus advanced (T3/T4) stage discrimination and the crucial T4 versus non-T4 decision for treatment planning. Each architecture was rigorously evaluated using LaryngealCT, the first curated, ROI-focused laryngeal cancer CT dataset, aggregating 1,029 scans from six major collections from The Cancer Imaging Archive (TCIA)[20–26]. The dataset provides 1 mm isotropic volumes of interest, comprehensive metadata, and expert-validated data processing, ensuring standardisation across tumour stages and anatomical subsites.

Our benchmarking pipeline not only quantifies model performance but also integrates clinical metadata and detailed explainability analyses, contextualising predictions within laryngeal anatomy. By making the dataset and validated benchmarking workflows openly available, this work establishes robust, reproducible standards to drive the next generation of AI tools for laryngeal oncology, promoting transparency, generalisability, and clinical relevance in future research.

# Results

## Dataset Characteristics

The final curated dataset comprised 1,029 laryngeal cancer CT volumes aggregated from six publicly available TCIA datasets (see Table 2). The cohort spanned all tumour T-stages, including 40 Tis, 312 T1, 286 T2, 307 T3, and 84 T4 cases (Fig. 1a). Clinical metadata such as age, sex, tumour subsite, and treatment type were collated and indexed, providing a resource for supervised learning and clinical correlation.

**Table 2 | T-stage-wise distribution of laryngeal cancer cases across TCIA datasets**

| Sl. No. | Name of Dataset | Total no. of laryngeal cancer cases | No. of laryngeal cancer cases per T-stage | | | | |
|---|---|---|---|---|---|---|---|
| | | | Tis | T1 | T2 | T3 | T4 |
| **1** | RADCURE | 867 | 40 | 281 | 253 | 233 | 60 |
| **2** | Head-Neck-PET-CT | 44 | 00 | 03 | 10 | 28 | 03 |
| **3** | HEAD-NECK-RADIOMICS-HN1 | 49 | 00 | 21 | 04 | 11 | 13 |
| **4** | HNSCC | 19 | 00 | 00 | 01 | 16 | 02 |
| **5** | HNSCC-3DCT-RT | 04 | 00 | 01 | 00 | 02 | 01 |
| **6** | QIN-HEADNECK | 46 | 00 | 06 | 18 | 17 | 05 |
| **Total** | | **1029** | **40** | **312** | **286** | **307** | **84** |

To support transparency and reproducibility, all preprocessing resources are available on our GitHub repository (see Data Availability section). This includes dataset download instructions, a step-by-step workflow, cropping parameters, and code snippets to automatically generate laryngeal volumes and structured metadata from TCIA series.

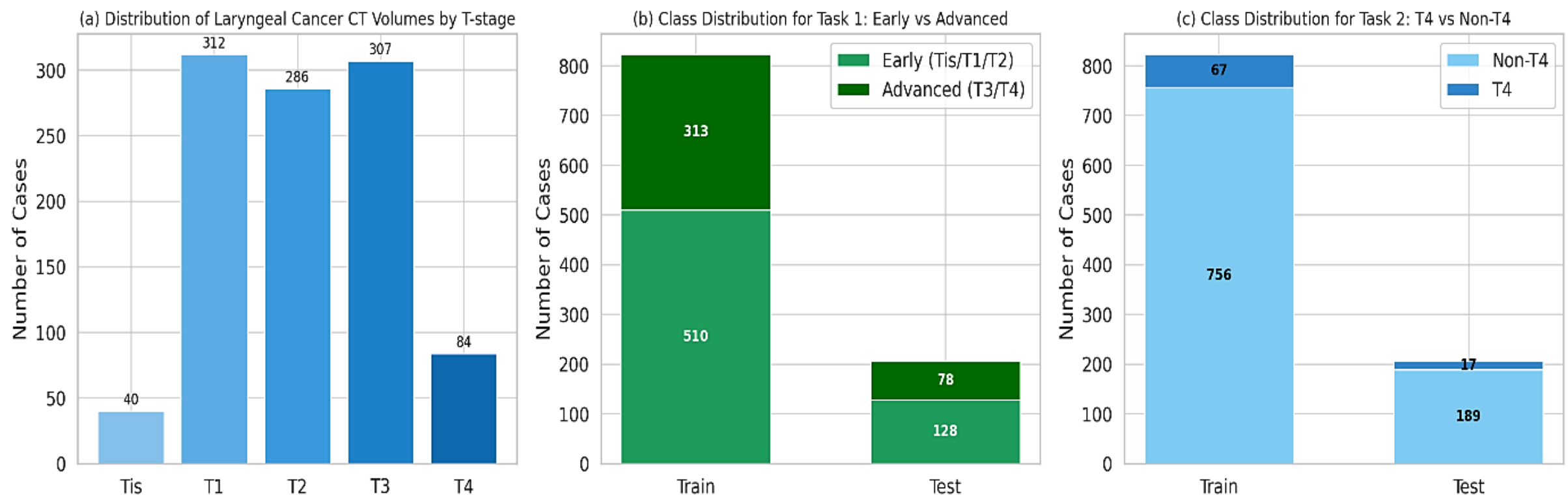

**Fig. 1 | Class distribution of the curated LaryngealCT dataset**. (a) Distribution of cases across individual T-stages (Tis, T1, T2, T3, T4), highlighting coverage across the clinical spectrum of disease. (b) Aggregated distribution for Task 1 (early: Tis/T1/T2 vs. advanced: T3/T4), showing balanced representation in training and test subsets. (c) Distribution for Task 2 (T4 vs. non-T4), demonstrating the class imbalance that models must address.

For model development, the dataset was divided into 823 training and 206 independent test cases, stratified across tumour stages and source datasets to minimise sampling bias. For the first classification task (early [Tis/T1/T2] vs. advanced [T3/T4]), the training set included 510 early and 313 advanced cases, while the test set comprised 128 early and 78 advanced cases (Fig. 1b). For the second classification task (T4 vs. non-T4), the training set contained 756 non-T4 and 67 T4 cases, with 189 non-T4 and 17 T4 cases in the test set (Fig. 1c). This distribution highlights a substantial class imbalance, particularly for the T4 category, which is clinically important to account for in benchmarking analyses.

## Validation of Cropping

To evaluate both the clinical fidelity and the reproducibility of the cropping pipeline, we performed complementary expert review and quantitative analysis.

First, a representative subset of 100 cropped CT volumes was independently reviewed by a senior radiologist and a senior head and neck surgeon. Ratings were assigned using a 5-point Likert scale (1 = Poor, 5 = Excellent). Inter-rater reliability was quantified using weighted Cohen's Kappa statistics and raw agreement percentages with 95% confidence intervals. Perfect agreement ($\kappa$ = 1.000) was observed for anatomical coverage, underscoring the robustness of the cropping protocol in consistently isolating the target region of interest. For the remaining criteria, Kappa values ranged from 0.901 to 0.936, representing almost perfect agreement[30]. Raw agreement exceeded 87% across all dimensions, with image quality achieving 91% concordance despite marginal variability in scale usage (see Table 3).

**Table 3 | Inter-rater agreement metrics for expert evaluation**

| Criterion | Weighted Kappa (95% CI) | Raw Agreement% (95% CI) |
| --- | --- | --- |
| **Anatomical Coverage** | 1.00<br>(Perfect agreement, not estimable*) | 100<br>(Perfect agreement, not estimable*) |
| **Image Quality** | 0.904<br>(0.840, 0.957) | 91<br>(85.0, 96.0) |
| **Segmentation Feasibility** | 0.901<br>(0.823, 0.942) | 87<br>(80.0, 93.0) |
| **Classification Feasibility** | 0.936<br>(0.854, 0.976) | 93<br>(88.0, 97.0) |

*For criteria with perfect agreement, confidence intervals are not estimable due to zero variance.

Quantitative validation was performed across the full dataset (n = 1,029) by comparing parameter-search crops with their manually derived counterparts. For 1,025 cases (99.6%), automated crops exhibited near-perfect concordance with manual volumes, achieving a mean Dice similarity coefficient of 0.995 (SD 0.0013) and mean Intersection-over-Union of 0.990 (SD 0.0026); even the lowest Dice score exceeded 0.985. Supporting voxel-wise error metrics also confirmed fidelity (MAE = 0.99, MSE = 1.9). Only four cases (<0.4%) could not be assessed quantitatively because their source CTs had substantially larger slice counts and field-of-view dimensions, leading to bounding boxes that extended outside the standardised reconstruction grid. These were automatically clamped to fit within bounds, yielding reproducible but slightly smaller crops than the manual references. Collectively, these results validate the reproducibility and geometric accuracy of the preprocessing pipeline, establishing confidence that automated crops are essentially indistinguishable from expert manual crops for downstream analysis. Detailed case-wise metrics (CSV and JSON files) and distributional summaries (boxplots) are made publicly available on our GitHub repository (see Data Availability section).

Together, the high expert agreement and strong quantitative concordance confirm that the proposed preprocessing pipeline produces clinically valid and reproducible laryngeal crops suitable for downstream segmentation and classification tasks.

## DL Benchmarking Performance

We benchmarked six 3D convolutional neural network (CNN) architectures: a custom lightweight 5-layer 3D CNN (see Methods section for details), ResNet18, ResNet50, ResNet101[27], DenseNet121[28], and a transfer learning variant of ResNet50 pretrained on MedicalNet[29]. These models span increasing network depth and connectivity, enabling assessment of trade-offs between representational capacity, generalisation, and computational cost.

### Cross-Validation Performance of DL Models

The benchmarking of six deep learning architectures under 5-fold cross-validation revealed distinct performance trends across the two clinically motivated classification tasks. Table 4 summarises the mean cross-validation performances of the DL models after calibration (reported as Mean±SD).

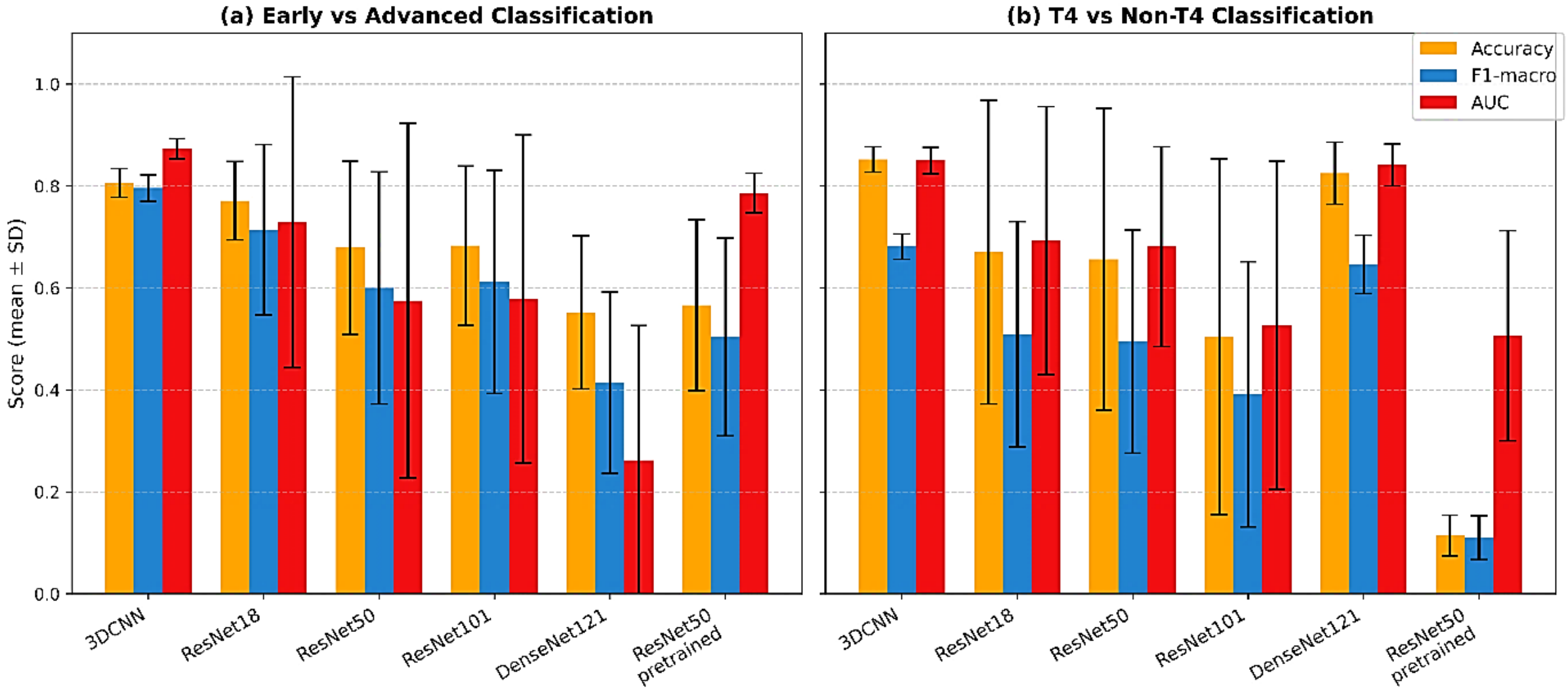

**Fig. 2 | Cross-validation performance of six DL architectures on the curated LaryngealCT dataset.** (a) Early (Tis/T1/T2) vs. Advanced (T3/T4) staging. (b) T4 vs. non-T4 classification. Bars show mean Accuracy, F1-macro, and AU-ROC across 5 folds, with error bars representing standard deviation. The 3D CNN consistently achieved strong and stable performance, particularly in the T4 classification task, while deeper residual networks showed greater variability across folds.

### Early versus advanced classification (Tis/T1/T2 vs. T3/T4):

Among the DL models, the custom 3D CNN achieved the most stable and generalisable performance compared to the deeper architectures, achieving the highest balanced accuracy (0.798 ± 0.019), F1-macro (0.796 ± 0.026), and Area Under the Curve (AU-ROC) (0.873 ± 0.020). Its sensitivity (0.767 ± 0.020) was notably higher than that of all ResNet variants and DenseNet121, indicating superior ability to detect advanced-stage cases. ResNet18 ranked second, with moderate F1-macro (0.714 ± 0.167) and high specificity (0.876 ± 0.069) but suffered from reduced sensitivity (0.599 ± 0.305), suggesting a bias toward early-stage classification. Deeper residual networks (ResNet50/101) and DenseNet121 demonstrated progressive performance degradation, reflected in reduced AU-ROCs (0.575–0.578) and unstable sensitivity estimates (0.492–0.618), likely attributable to overfitting and limited sample size. Transfer learning with MedicalNet-pretrained ResNet50 yielded modest improvements in calibration (AU-ROC = 0.786 ± 0.039), but its sensitivity–specificity balance remained suboptimal (0.717 vs. 0.475), limiting clinical utility.

**Table 4 | Cross Validation Metrics for DL models**

| Model | Accuracy | Bal.Accuracy | F1_macro | AU-ROC | Sensitivity | Specificity |
|---|---|---|---|---|---|---|
| **Early vs Advanced Classification** | | | | | | |
| **3DCNN** | **0.806±0.028** | **0.798±0.019** | **0.796±0.026** | **0.873±0.020** | **0.767±0.020** | 0.829±0.055 |
| **ResNet18** | 0.771±0.077 | 0.738±0.121 | 0.714±0.167 | 0.729±0.285 | 0.599±0.305 | **0.876±0.069** |
| **ResNet50** | 0.679±0.170 | 0.668±0.141 | 0.600±0.228 | 0.575±0.348 | 0.618±0.343 | 0.718±0.370 |
| **ResNet101** | 0.683±0.156 | 0.646±0.194 | 0.612±0.219 | 0.578±0.322 | 0.492±0.384 | 0.800±0.131 |
| **DenseNet121** | 0.552±0.150 | 0.554±0.108 | 0.414±0.178 | 0.262±0.264 | 0.561±0.464 | 0.547±0.457 |
| **ResNet50 pretrained** | 0.566±0.168 | 0.596±0.125 | 0.504±0.194 | 0.786±0.039 | 0.717±0.317 | 0.475±0.387 |
| **T4 Classification** | | | | | | |
| **3DCNN** | **0.852±0.025** | **0.798±0.028** | **0.681±0.025** | **0.850±0.026** | 0.733±0.070 | **0.863±0.030** |
| **ResNet18** | 0.670±0.298 | 0.659±0.085 | 0.509±0.221 | 0.693±0.263 | 0.645±0.181 | 0.673±0.339 |
| **ResNet50** | 0.656±0.296 | 0.638±0.078 | 0.495±0.219 | 0.681±0.196 | 0.615±0.193 | 0.660±0.338 |
| **ResNet101** | 0.504±0.349 | 0.628±0.105 | 0.391±0.260 | 0.527±0.322 | 0.776±0.199 | 0.480±0.397 |
| **DenseNet121** | 0.825±0.061 | 0.748±0.020 | 0.646±0.057 | 0.841±0.041 | 0.656±0.040 | 0.840±0.069 |
| **ResNet50 pretrained** | 0.114±0.040 | 0.491±0.011 | 0.110±0.043 | 0.506±0.206 | **0.941±0.073** | 0.041±0.050 |

All metrics reported are calibrated at optimal threshold, averaged across folds, and reported as Mean±SD; best metrics highlighted in bold. Bal.Accuracy = Balanced Accuracy; AU-ROC = Area Under the Receiver Operating Characteristics Curve.

### T4 versus non-T4 classification:

In the more clinically critical T4 detection task, the custom 3D CNN again achieved the strongest overall performance, with balanced accuracy of 0.798 ± 0.028, F1-macro of 0.681 ± 0.025, and AU-ROC of 0.850 ± 0.026. The model also demonstrated robust sensitivity (0.733 ± 0.070) and specificity (0.863 ± 0.030), indicating reliable discrimination despite class imbalance. DenseNet121 emerged as a competitive alternative (AU-ROC = 0.841 ± 0.041; F1-macro = 0.646 ± 0.057), though its sensitivity was lower (0.656 ± 0.040), potentially limiting detection of true T4 cases. By contrast, ResNet18 and ResNet50 exhibited variable and less stable performance (F1-macro 0.495–0.509, AU-ROC 0.681–0.693), while ResNet101 was the weakest residual variant (F1-macro = 0.391 ± 0.260). Surprisingly, transfer learning with pretrained ResNet50 collapsed in this setting, yielding very low overall accuracy (0.114 ± 0.040) and balanced accuracy (0.491 ± 0.011) despite high sensitivity (0.941 ± 0.073). This indicates over-sensitivity at the expense of specificity (0.041 ± 0.050), leading to severe misclassification of non-T4 cases. Fig. 2 illustrates the comparison of major metrics (Accuracy, F1-macro and AU-ROC) across the DL models for both classification tasks.

Across both tasks, the lightweight 3D CNN provided the most stable and generalisable performance, outperforming deeper state-of-the-art architectures under limited-sample conditions. The results emphasise that, in domain-specific imaging tasks with inherent class imbalance, architectural simplicity coupled with task-tailored design may offer superior clinical utility compared to deeper networks or off-the-shelf transfer learning.

## Independent Test Performance

On the independent test cohort (n = 206), all models maintained broadly consistent performance with cross-validation estimates, though differences in sensitivity–specificity balance became more evident.

### Early versus advanced classification (Tis/T1/T2 vs. T3/T4):

The overall test metrics are summarised in Table 5, which reports accuracy, balanced accuracy, AU-ROC, F1-macro, and other global indicators. In addition, per-class precision, recall, and F1-scores are detailed in Table 6, enabling a more granular view of performance under class imbalance. Fig. 3 presents ROC overlays comparing validation and independent test performance across all models, demonstrating close agreement between mean validation AU-ROCs and test AU-ROCs, with only minor deviations. Unless otherwise specified, overall test-set metrics in Tables 5 and 7 are reported at the default decision threshold of 0.5, whereas Tables 6 and 8 provide both raw (default-threshold) and calibrated (temperature-scaled, threshold-optimised) per-class metrics for each model.

**Table 5 | Overall test metrics for early vs. advanced classification across six DL models**

| Model | 3DCNN | ResNet18 | ResNet50 | ResNet101 | DenseNet121 | ResNet50 Pre-trained |
|---|---|---|---|---|---|---|
| **Accuracy** | **0.854** (0.806-0.903) | 0.801 (0.742-0.854) | 0.835 (0.782-0.879) | 0.830 (0.777-0.879) | 0.786 (0.728-0.840) | 0.767 (0.704-0.825) |
| **Bal. Accuracy** | **0.833** (0.779-0.882) | 0.772 (0.709-0.832) | 0.805 (0.745-0.860) | 0.818 (0.766-0.871) | 0.761 (0.696-0.820) | 0.747 (0.688-0.808) |
| **F1-macro** | **0.841** (0.785-0.890) | 0.780 (0.717-0.834) | 0.816 (0.758-0.869) | 0.819 (0.766-0.870) | 0.767 (0.695-0.827) | 0.750 (0.684-0.808) |
| **Sensitivity** | 0.744 (0.646-0.839) | 0.654 (0.544-0.753) | 0.679 (0.577-0.772) | **0.769** (0.671-0.862) | 0.654 (0.543-0.756) | 0.667 (0.557-0.760) |
| **Specificity** | 0.922 (0.870-0.962) | 0.891 (0.832-0.938) | **0.930** (0.881-0.970) | 0.867 (0.811-0.923) | 0.867 (0.799-0.922) | 0.828 (0.764-0.892) |
| **AU-ROC** | **0.877** (0.828-0.928) | 0.867 (0.809-0.918) | 0.871 (0.818-0.919) | 0.865 (0.809-0.912) | 0.865 (0.810-0.912) | 0.813 (0.748-0.868) |
| **PR-AUC** | **0.844** (0.763-0.906) | **0.844** (0.773-0.904) | 0.818 (0.733-0.900) | 0.819 (0.726-0.895) | 0.837 (0.761-0.896) | 0.709 (0.603-0.815) |
| **F1-weighted** | **0.852** | 0.801 | 0.815 | 0.834 | 0.744 | 0.751 |
| **MCC** | **0.686** | 0.568 | 0.644 | 0.638 | 0.537 | 0.500 |
| **Brier Score** | 0.181 | **0.140** | 0.150 | 0.162 | 0.160 | 0.188 |
| **PPV** | 0.853 | 0.785 | **0.855** | 0.779 | 0.750 | 0.703 |
| **NPV** | **0.855** | 0.809 | 0.826 | 0.860 | 0.804 | 0.803 |

All metrics are reported at default threshold = 0.5; 95% confidence intervals in parentheses. Bal.Accuracy=Balanced Accuracy; AU-ROC=Area Under the Receiver Operating Characteristics Curve; PR-AUC=Area Under the Precision-Recall Curve; MCC=Matthews Correlation Coefficient; PPV=Positive Predictive Value; NPV=Negative Predictive Value. Best metrics highlighted in bold.

The custom 3D CNN achieved the strongest overall balance, with accuracy of 0.854 (95% CI: 0.806–0.903), balanced accuracy of 0.833 (0.779–0.882), and F1-macro of 0.841 (0.785–0.890). Sensitivity was 0.744 (0.646–0.839) and specificity 0.922 (0.870–0.962), indicating reliable discrimination across both classes.

ResNet18 achieved accuracy of 0.801 (0.742–0.854) and AU-ROC of 0.867 (0.809–0.918), but sensitivity was lower at 0.654 (0.544–0.753), reflecting bias toward early-stage classification despite high specificity (0.891). ResNet50 and ResNet101 performed comparably, with AU-ROC values of 0.871 (0.818–0.919) and 0.865 (0.809–0.912), respectively, but showed moderate sensitivity (0.679 and 0.769) and balanced accuracy (0.805 and 0.818). DenseNet121 achieved accuracy of 0.786 (0.728–0.840) and AU-ROC of 0.865 (0.810–0.912), but sensitivity remained limited (0.654). The pretrained ResNet50 variant produced the weakest results, with accuracy of 0.767 (0.704–0.825), balanced accuracy of 0.747 (0.688–0.808), and AU-ROC of 0.813 (0.748–0.868).

**Table 6 | Per-class precision, recall, and F1-scores for early vs. advanced classification**

| Model | | Per-class Metrics | | | | | | Confusion Matrix | | | |
|---|---|---|---|---|---|---|---|---|---|---|---|
| | | Class 0 - Early Stage (Support=128) | | | Class 1 – Advanced Stage (Support=78) | | | | | | |
| | | Precision | Recall | F1 | Precision | Recall | F1 | TN | FP | FN | TP |
| **3D CNN** | Raw (95% CI) | 0.855 (0.794-0.909) | 0.922 (0.872-0.965) | **0.887** (0.841-0.925) | 0.853 (0.768-0.928) | 0.744 (0.641-0.842) | **0.775** (0.715-0.862) | 118 | 10 | 20 | 58 |
| | Cal. | **0.864** | **0.891** | **0.877** | **0.811** | **0.769** | **0.789** | 114 | 14 | 18 | 60 |
| | | | | | | | | t_opt:0.484, T:0.889 | | | |
| **ResNet18** | Raw (95% CI) | 0.809 (0.758-0.883) | 0.891 (0.837-0.942) | 0.848 (0.811-0.897) | 0.785 (0.699-0.887) | 0.654 (0.583-0.791) | 0.713 (0.648-0.818) | 114 | 14 | 27 | 51 |
| | Cal. | 0.809 | 0.891 | 0.848 | 0.785 | 0.654 | 0.713 | 114 | 14 | 27 | 51 |
| | | | | | | | | t_opt:0.490, T:-93.118 | | | |
| **ResNet50** | Raw (95% CI) | 0.826 (0.764-0.885) | **0.930** (0.881-0.969) | 0.875 (0.832-0.914) | **0.855** (0.759-0.933) | 0.679 (0.576-0.776) | 0.757 (0.667-0.831) | 119 | 9 | 25 | 53 |
| | Cal. | 0.844 | 0.891 | 0.867 | 0.803 | 0.731 | 0.765 | 114 | 14 | 21 | 57 |
| | | | | | | | | t_opt:0.438, T:-152.123 | | | |
| **ResNet 101** | Raw (95% CI) | **0.860** (0.804-0.918) | 0.867 (0.803-0.924) | 0.864 (0.817-0.906) | 0.779 (0.686-0.862) | **0.769** (0.675-0.861) | 0.774 (0.701-0.840) | 111 | 17 | 18 | 60 |
| | Cal. | 0.857 | 0.891 | 0.874 | 0.808 | 0.756 | 0.781 | 114 | 14 | 19 | 59 |
| | | | | | | | | t_opt:0.509, T:-65.797 | | | |
| **Dense Net121** | Raw (95% CI) | 0.804 (0.737-0.867) | 0.867 (0.810-0.922) | 0.835 (0.784-0.879) | 0.750 (0.642-0.855) | 0.654 (0.554-0.753) | 0.669 (0.605-0.778) | 111 | 17 | 27 | 51 |
| | Cal. | 0.787 | 0.836 | 0.811 | 0.700 | 0.628 | 0.662 | 107 | 21 | 29 | 49 |
| | | | | | | | | t_opt:0.516, T:-237.513 | | | |
| **ResNet50 pre-trained** | Raw (95% CI) | 0.803 (0.735-0.873) | 0.828 (0.754-0.891) | 0.815 (0.762-0.863) | 0.703 (0.600-0.797) | 0.667 (0.554-0.765) | 0.684 (0.597-0.756) | 106 | 22 | 26 | 52 |
| | Cal. | 0.715 | 0.883 | 0.790 | 0.688 | 0.423 | 0.524 | 113 | 15 | 45 | 33 |
| | | | | | | | | t_opt:0.503, T:45.828 | | | |

Raw metrics at default threshold = 0.5; calibrated values reflect optimal thresholds (t_opt) after temperature (T) scaling. Best metrics highlighted in bold.

Per-class metrics (Table 6) confirmed these trends. The 3D CNN identified early-stage cases with precision 0.855, recall 0.922, F1 0.887, and advanced cases with precision 0.853, recall 0.744, F1 0.775 at the default threshold. Calibration improved advanced-stage recall to 0.769, yielding a more balanced profile. ResNet18 and ResNet50 achieved similar AU-ROC values but lower advanced-stage recall (0.654–0.679). ResNet101 showed competitive AU-ROC but unstable sensitivity after calibration. DenseNet121 and pretrained ResNet50 underperformed relative to lighter models. Overall, the 3D CNN provided the most consistently strong performance across global and per-class metrics, but sensitivity for advanced disease remained limited, underscoring the challenge of balanced detection.

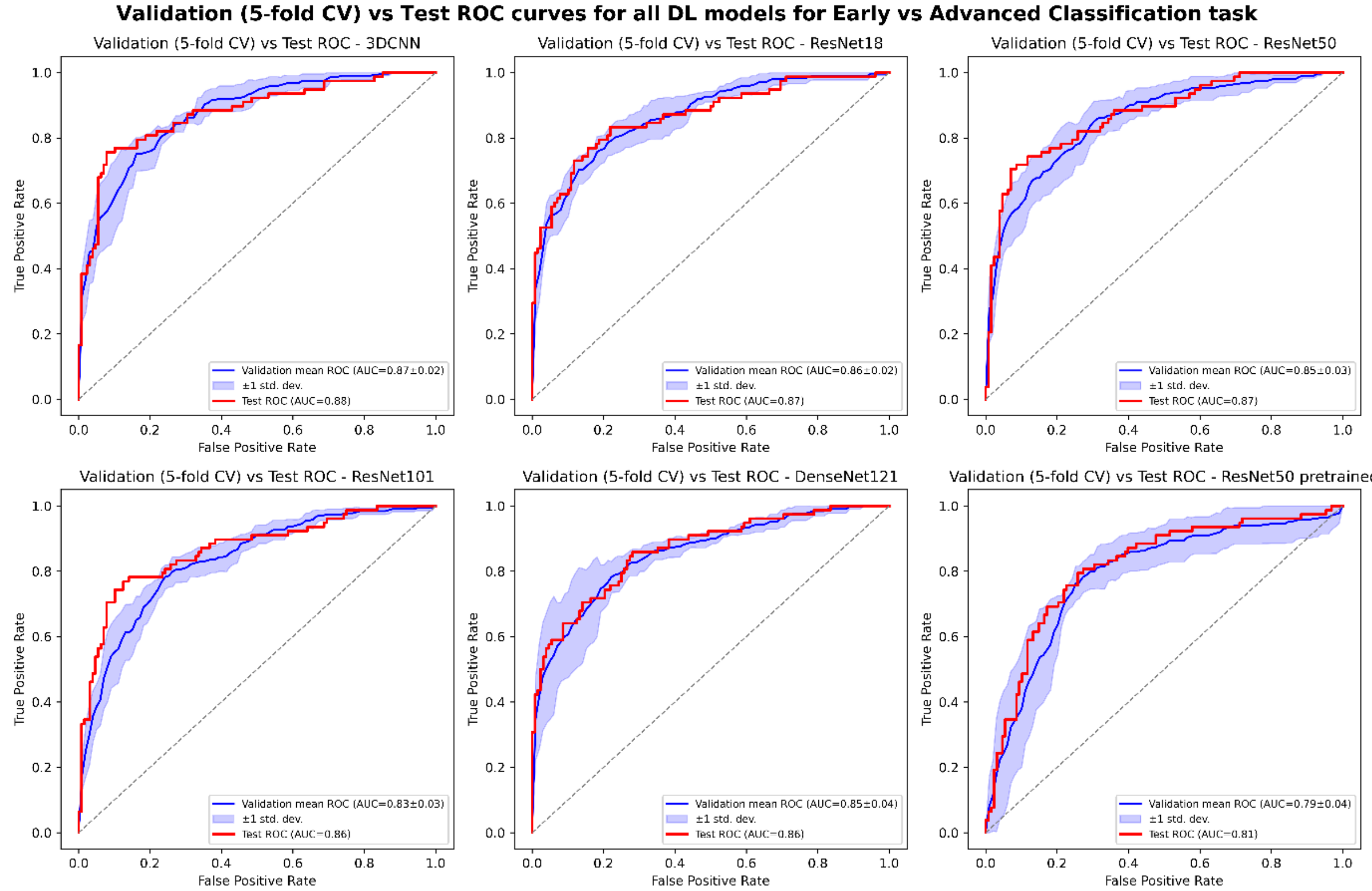

**Fig. 3 | Overlays comparing mean validation ROC ±1 SD (shaded) and independent test ROC (solid) for early vs. advanced classification across all six deep learning models.**

### T4 versus non-T4 classification:

The overall test outcomes are reported in Table 7, while per-class metrics are provided in Table 8 to highlight the disparity between non-T4 and T4 case recognition. Fig. 4 presents ROC overlays comparing validation and test performance for T4 versus non-T4 classification. Test AU-ROCs were generally consistent with mean validation AU-ROCs, and the overall discrimination appeared strong (AU-ROC ≥0.82 for most models), but sensitivity for T4 cases was consistently low. This task underscored the difficulty of detecting a clinically rare minority class.

**Table 7 | Overall test metrics for T4 vs. non-T4 classification across six DL models**

| Model | 3DCNN | ResNet18 | ResNet50 | ResNet101 | DenseNet121 | ResNet50 Pre-trained |
|---|---|---|---|---|---|---|
| **Accuracy** | 0.854 (0.805-0.898) | 0.908 (0.869-0.942) | **0.922** (0.883-0.956) | 0.913 (0.874-0.951) | 0.913 (0.869-0.947) | 0.917 (0.879-0.951) |
| **Bal. Accuracy** | 0.653 (0.537-0.782) | 0.602 (0.510-0.719) | 0.610 (0.521-0.723) | **0.685** (0.573-0.812) | 0.631 (0.526-0.739) | 0.500 (0.491-0.598) |
| **F1-macro** | 0.618 (0.517-0.710) | 0.623 (0.507-0.732) | 0.646 (0.516-0.776) | **0.695** (0.573-0.805) | 0.655 (0.531-0.770) | 0.478 (0.468-0.488) |
| **Sensitivity** | 0.411 (0.182-0.667) | 0.235 (0.059-0.467) | 0.235 (0.045-0.455) | **0.412** (0.190-0.667) | 0.294 (0.077-0.500) | 0.000 (0.000-0.195) |
| **Specificity** | 0.894 (0.853-0.938) | 0.968 (0.942-0.990) | **0.984** (0.963-1.000) | 0.958 (0.927-0.984) | 0.968 (0.942-0.990) | 1.000 (0.981-1.000) |
| **AU-ROC** | 0.873 (0.816-0.925) | 0.827 (0.723-0.916) | 0.820 (0.723-0.904) | 0.859 (0.771-0.930) | 0.879 (0.812-0.935) | 0.880 (0.790-0.958) |
| **PR-AUC** | 0.380 (0.200-0.601) | 0.445 (0.230-0.651) | 0.416 (0.200-0.635) | 0.446 (0.217-0.656) | **0.464** (0.239-0.665) | 0.478 (0.262-0.722) |
| **F1-weighted** | 0.869 | 0.904 | **0.907** | 0.903 | 0.904 | 0.910 |
| **MCC** | 0.249 | 0.261 | 0.333 | **0.391** | 0.321 | 0.00 |
| **Brier Score** | 0.163 | 0.076 | 0.080 | 0.085 | **0.067** | 0.096 |

| PPV | 0.259 | 0.400 | **0.571** | 0.467 | 0.455 | 0.000 |
|---|---|---|---|---|---|---|
| NPV | 0.944 | 0.934 | 0.935 | **0.948** | 0.938 | 1.000 |

All metrics are reported at default threshold = 0.5; 95% confidence intervals in parentheses. Bal.Accuracy=Balanced Accuracy; AU-ROC=Area Under the Receiver Operating Characteristics Curve; PR-AUC=Area Under the Precision-Recall Curve; MCC=Matthews Correlation Coefficient; PPV=Positive Predictive Value; NPV=Negative Predictive Value. Best metrics highlighted in bold.

The 3D CNN achieved accuracy of 0.854 (0.805–0.898), balanced accuracy of 0.653 (0.537–0.782), and AU-ROC of 0.873 (0.816–0.925). However, sensitivity for T4 cases was only 0.411 (0.182–0.667), despite high specificity (0.894). ResNet18 achieved higher overall accuracy (0.908) and AU-ROC (0.827), but sensitivity was even lower (0.235). ResNet50 achieved the highest accuracy (0.922) and specificity (0.984), but T4 sensitivity remained 0.235. ResNet101 achieved accuracy of 0.913, balanced accuracy of 0.685, and AU-ROC of 0.859, with sensitivity 0.412, the highest among the residual networks. DenseNet121 achieved accuracy of 0.913, AU-ROC of 0.879, and sensitivity 0.294. The pretrained ResNet50 collapsed completely, with T4 sensitivity of 0.000 (0.000–0.195) despite AU-ROC of 0.880, reflecting extreme bias toward the majority class.

**Table 8 | Per-class precision, recall, and F1-scores for T4 vs. non-T4 classification**

| Model | | Per-class Metrics | | | | | | Confusion Matrix | | | |
|---|---|---|---|---|---|---|---|---|---|---|---|
| | | Class 0 – non-T4 (Support:189) | | | Class 1 – T4 (Support:17) | | | | | | |
| | | Precision | Recall | F1 | Precision | Recall | F1 | TN | FP | FN | TP |
| 3D CNN | Raw (95% CI) | **0.944** (0.909-0.973) | 0.894 (0.847-0.935) | 0.918 (0.888-0.944) | 0.259 (0.100-0.433) | **0.412** (0.176-0.667) | 0.318 (0.129-0.488) | 169 | 20 | 10 | 7 |
| | Cal. | 0.944 | **0.884** | **0.913** | 0.241 | 0.412 | 0.304 | 167 | 22 | 10 | 7 |
| | | | | | | | | t_opt:0.476, T=0.234 | | | |
| ResNet 18 | Raw (95% CI) | 0.934 (0.896-0.965) | 0.968 (0.943-0.990) | 0.951 (0.926-0.971) | 0.400 (0.110-0.750) | 0.235 (0.059-0.455) | 0.296 (0.077-0.522) | 183 | 6 | 13 | 4 |
| | Cal. | 0.957 | 0.825 | 0.886 | 0.233 | 0.588 | 0.333 | 156 | 33 | 7 | 10 |
| | | | | | | | | t_opt:0.238, T=-79.513 | | | |
| ResNet 50 | Raw (95% CI) | 0.935 (0.899-0.970) | **0.984** (0.963-1.000) | **0.959** (0.937-0.977) | **0.571** (0.098-1.000) | 0.235 (0.059-0.438) | 0.333 (0.091-0.563) | 186 | 3 | 13 | 4 |
| | Cal. | 0.952 | 0.836 | 0.890 | 0.225 | 0.529 | 0.315 | 158 | 31 | 8 | 9 |
| | | | | | | | | t_opt:0.270, T:-19.084 | | | |
| ResNet 101 | Raw (95% CI) | 0.948 (0.915-0.978) | 0.958 (0.930-0.984) | 0.953 (0.930-0.974) | 0.467 (0.214-0.750) | **0.412** (0.167-0.650) | **0.438** (0.214-0.632) | 181 | 8 | 10 | 7 |
| | Cal. | **0.969** | 0.825 | 0.891 | 0.267 | **0.706** | **0.387** | 156 | 33 | 5 | 12 |
| | | | | | | | | t_opt:0.340, T:-53.530 | | | |
| Dense net 121 | Raw (95% CI) | 0.938 (0.901-0.970) | 0.968 (0.942-0.990) | 0.953 (0.929-0.974) | 0.455 (0.167-0.800) | 0.294 (0.083-0.533) | 0.357 (0.095-0.563) | 183 | 6 | 12 | 5 |
| | Cal. | 0.954 | **0.884** | **0.918** | **0.290** | 0.529 | 0.375 | 167 | 22 | 8 | 9 |
| | | | | | | | | t_opt:0.264, T:0.846 | | | |
| ResNet 50 pretrained | Raw (95% CI) | 0.917 (0.873-0.951) | 1.000 (0.000-0.195) | 0.957 (0.935-0.978) | 0.000 (N/A) | 0.000 (0.000-0.195) | 0.000 (N/A) | 189 | 0 | 17 | 0 |
| | Cal. | 0.000 | 0.000 | 0.000 | 0.083 | 1.000 | 0.152 | 0 | 189 | 0 | 17 |
| | | | | | | | | t_opt:0.312, T: -197.361 | | | |

Raw metrics at default threshold = 0.5; calibrated values reflect optimal thresholds (t_opt) after temperature (T) scaling. Best metrics highlighted in bold.

Per-class metrics (Table 8) confirmed these limitations. For the 3D CNN, non-T4 cases were recognised with precision 0.944, recall 0.894, F1 0.918, but T4 cases achieved only precision 0.259, recall 0.412, F1 0.318. Calibration did not substantially improve minority class recall.

ResNet18 achieved non-T4 F1 of 0.951 but T4 F1 of 0.296 at default threshold, improving to 0.333 after calibration. ResNet50 achieved non-T4 F1 of 0.959 but T4 F1 of 0.333. ResNet101 showed the most balanced profile, with calibrated T4 recall of 0.706 and F1 of 0.387, though overall balanced accuracy remained modest. DenseNet121 achieved calibrated T4 F1 of 0.375, while pretrained ResNet50 failed to identify any T4 cases.

Specificity remained very high across all models (0.91–1.00), reflecting strong recognition of non-T4 cases, but sensitivity for T4 detection was limited (<0.42 at default threshold). ResNet101 showed the most promising calibrated recall (0.706), but overall performance remained insufficient for clinical decision support. These results highlight the central challenge of imbalanced datasets in oncology AI research.

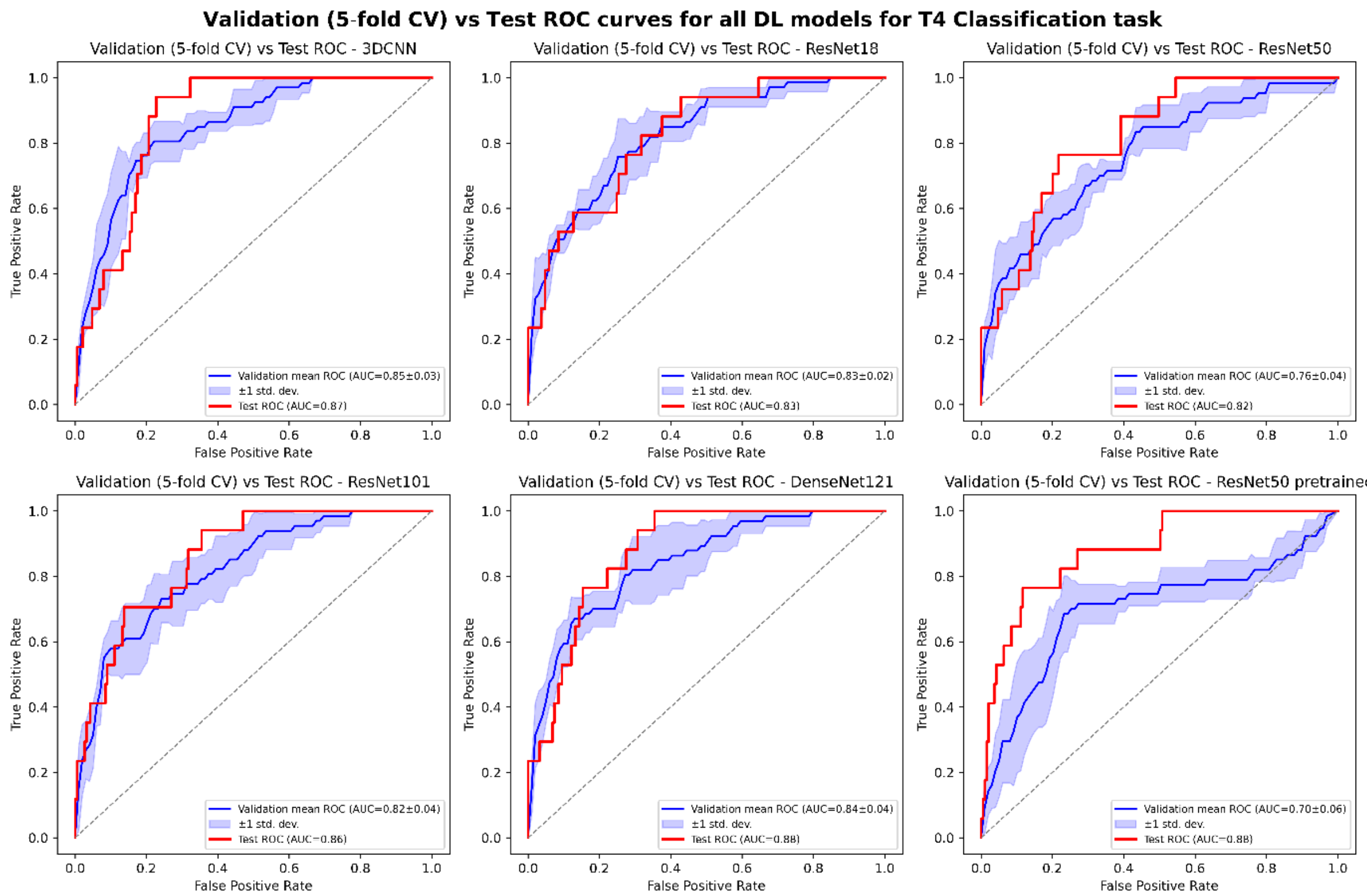

**Fig. 4 | Overlays comparing mean validation ROC ±1 SD (shaded) and independent test ROC (solid) for T4 vs. non-T4 classification across all six deep learning models.**

Taken together, these per-class analyses emphasise that while overall discrimination appears strong (AU-ROC ≥ 0.82 for most models), performance is disproportionately driven by the majority class. The inability to achieve robust recall for T4 disease highlights the central challenge of imbalanced datasets in oncology AI research, with direct implications for clinical adoption, where false negatives can result in undertreatment of advanced disease.

## Comparative Analysis and Clinical Implications

The comparative benchmarking highlights two central observations. First, lightweight architectures tailored to the laryngeal region, such as the custom 3D CNN, consistently outperformed deeper residual and densely connected networks across both classification tasks. This suggests that architectural simplicity, combined with careful preprocessing and calibration, may be more effective than depth in mid-sized, domain-specific datasets. Second, the clinical utility of these models depends not only on overall discrimination (AU-ROC) but also on their sensitivity to advanced disease. Across both tasks, high specificity was more easily achieved than balanced sensitivity, particularly in the T4 classification setting, where the number of positive cases was limited. From a clinical perspective, under-sensitivity risks missed diagnoses of thyroid cartilage invasion, which may have profound consequences for surgical decision-making.

Statistical comparisons reinforced these patterns. DeLong's test revealed significant differences ($p < 0.05$) in AU-ROC between all the models for the early vs advanced task, while differences in the T4 task were more modest ($p = 0.01$–$0.09$). Fig. 5 shows the heatmaps representing the p-values for the Delong tests conducted for both the classification tasks. Wilcoxon signed-rank tests across cross-validation folds showed overlapping distributions, confirming that superiority is relative rather than absolute. Also, ROC overlays comparing the mean validation curve (±1 SD) with the independent test performance (Figures 3–4) highlighted the stability of most of the models, with mean validation AU-ROCs closely matched the test AU-ROCs for the early vs advanced classification task. In contrast, the overlays showed greater deviation between validation and test, reflecting reduced generalisability for the T4 classification task compared to the first task. All metrics .csv files, plots, .json files, and confusion matrices are shared on our GitHub repository (see Data Availability section) for more details.

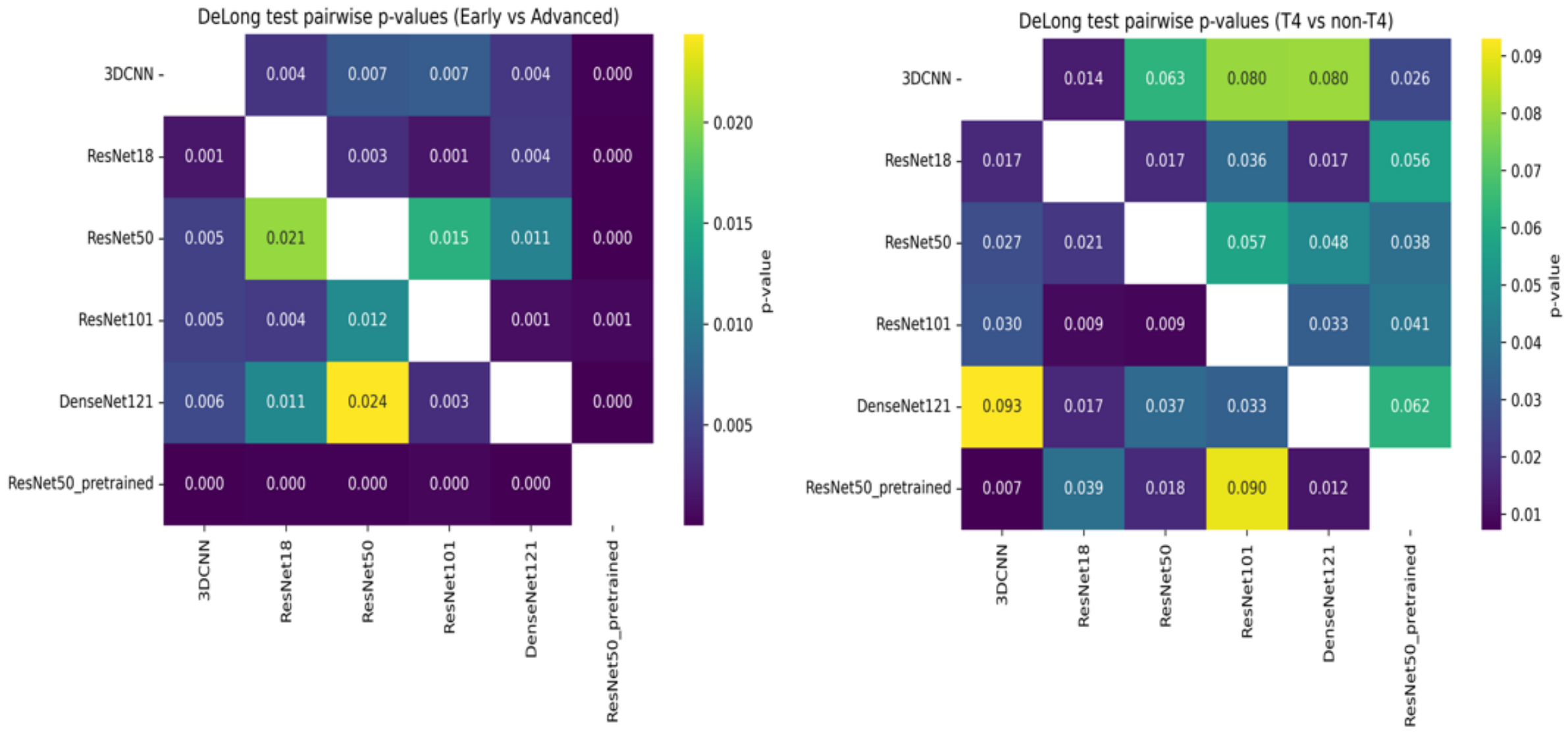

**Fig. 5 | Statistical comparison of model performances using heatmaps of pairwise DeLong test p-values for AU-ROC comparisons.** Lower p-values (<0.05) indicate statistically significant performance differences.

Together, these results underline the translational potential of lightweight, task-adapted networks in clinically constrained domains such as laryngeal cancer staging. However, they also highlight a persistent gap. Despite strong AU-ROC values, current models fail to achieve clinically acceptable recall for T4 disease. Addressing this gap will likely require targeted augmentation, cost-sensitive training, or hybrid radiomics–deep learning strategies that explicitly prioritise minority class detection.

Importantly, as this study introduces a benchmark dataset rather than a deployable clinical model, the observed limitations in T4 sensitivity should be interpreted as a reflection of the task's inherent complexity. The benchmarking results serve to highlight the need for improved methods and provide a reproducible foundation for future work focused on enhancing T4 detection. By openly reporting these challenges, LaryngealCT enables the community to iteratively improve upon baseline models and explore novel architectures, augmentation strategies, and multi-modal fusion approaches.

## Model Explainability and Clinical Interpretability

To explore how the models focus on laryngeal anatomy, we conducted GradCAM++ visualisation, perturbation-based sensitivity testing and mask-restricted overlap analysis for the T4 classification task.

Representative GradCAM++ maps for six cases (three T4 and three non-T4) showed that both the 3D CNN and ResNet101 highlighted anatomically plausible peri-cartilage and paraglottic regions. Correctly classified T4 cases exhibited focal activations along the cartilage boundaries, whereas non-T4 cases showed broader diffuse attention. These overlays illustrate exploratory interpretability patterns rather than definitive causal localisation.

CAM-guided deletion and insertion curves demonstrated monotonic behaviour, with model confidence decreasing when top-ranked voxels were removed and recovering upon reinsertion. For ResNet101, true positives showed stronger probability changes (ΔAU-ROC ≈ 0.030 for deletion, 0.027 for insertion) than false positives (ΔAU-ROC ≈ 0.019). In the 3D CNN, false negatives exhibited the largest shifts (ΔAU-ROC ≈ 0.166 for deletion, 0.092 for insertion), indicating that misclassified T4 cases contained highly influential voxels. Mask-guided perturbations revealed near-zero or negative probability changes overall, suggesting limited causal contribution of thyroid cartilage voxels to predictions.

Spatial correspondence between Grad-CAM++ activations and expert cartilage masks was modest. For ResNet101, Dice coefficients averaged 0.08 for true positives and 0.05 for false positives, with enrichment ratios of 1.6 vs 1.3. For the 3D CNN, Dice values were similarly low (<0.1), with enrichment ratios ranging from 1.3 to 1.6 depending on outcome type. These results indicate that while activations intersected cartilage regions more in correctly classified cases, absolute overlap remained limited.

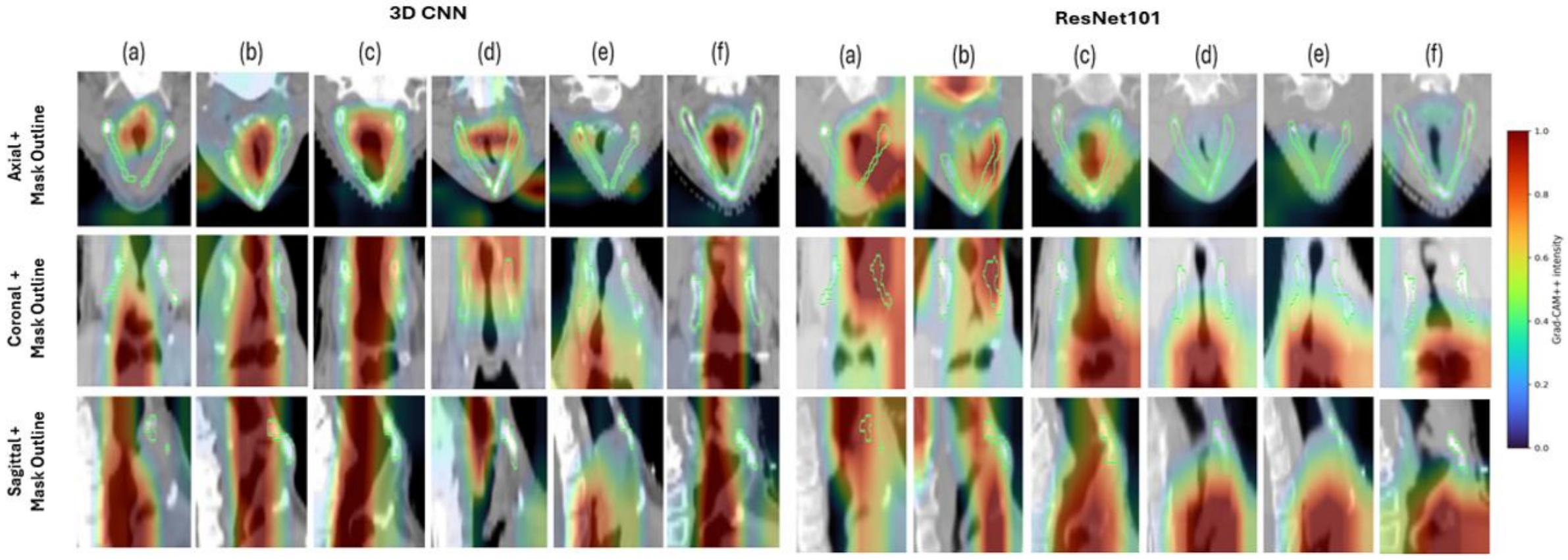

**Fig. 6 | 3D Grad-CAM++ visualisations overlaid with thyroid cartilage mask outlines for six representative cases.** Columns (a–f) correspond to the same patients for the 5-layer 3D CNN (left) and ResNet101 (right). Panels (a-b) show correctly classified T4 cases, panels (c-d) show non-T4 cases with (d) representing a false-positive prediction for the 3D CNN, while panels (e-f) show a false positive and a false negative case, respectively, for both models. Each panel displays axial, coronal, and sagittal slices, with thyroid cartilage contours overlaid in green. Both architectures highlight anatomically plausible peri-cartilage and paraglottic regions, though the extent and intensity of activations vary. These examples illustrate exploratory interpretability patterns rather than definitive causal localisation, underscoring the need for larger-scale validation.

Together, these analyses suggest that both architectures attend to anatomically plausible peri-cartilage regions, but causal coherence is weak. Perturbation curves confirm that salient voxels influence prediction but overlap with cartilage masks is modest. Accordingly, these findings are framed as exploratory and hypothesis-generating, underscoring the need for larger-scale validation and region-aware training strategies to strengthen interpretability.

Overall, the explainability analyses suggest that the classifier often attends to anatomically plausible peri-cartilage and surrounding soft-tissue regions, but relatively low CAM-mask overlap values and small sample of six representative cases mean that only tentative conclusions can be drawn about anatomical or causal coherence. The observed tendency to emphasize peri-cartilage context more than focal invasion cues is consistent with the residual false negatives and points to the need for future work on attention-guided, region-aware training to strengthen sensitivity to cartilage invasion.

# Discussion

This study introduces LaryngealCT, the first large-scale, curated, and clinically validated CT dataset for laryngeal cancer staging, accompanied by systematic benchmarking of six DL architectures. By focusing on an anatomically standardised volume of interest and validating its fidelity through expert review, we provide a clinically credible foundation for downstream AI pipelines. In this work, we benchmarked two clinically significant classification tasks: early versus advanced staging and T4 versus non-T4 discrimination, thereby addressing an unmet need in laryngeal oncology. While existing benchmarks such as MedSegBench[31] and FairMedFM[32] offer broad modality coverage and fairness evaluation, none capture the disease-specific nuances of laryngeal cancer or its CT-based staging tasks. LaryngealCT thus establishes a reproducible, anatomy-focused foundation for future model development and evaluation in this domain.

From a clinical standpoint, accurate identification of T4 tumours with thyroid cartilage invasion remains pivotal for treatment selection between organ-preserving strategies and total laryngectomy. Our benchmarking confirms that DL models can extract discriminative morphological cues, with AU-ROC values exceeding 0.82 across tasks, but sensitivity for T4 disease remained low on the independent test set. This under-sensitivity is clinically consequential as missed detections may lead to inappropriate larynx-preserving therapies and poorer survival outcomes, whereas high specificity supports confident exclusion of advanced disease in early-stage cases.

The custom 3D CNN achieved the best trade-off between sensitivity and specificity, with test accuracy of 0.854, balanced accuracy of 0.833, and AU-ROC of 0.877 in early vs advanced classification. In the T4 task, however, its sensitivity was only 0.411, highlighting the difficulty of minority class detection. ResNet101 achieved the most promising calibrated recall for T4 cases (0.706), but overall balanced accuracy remained modest. DenseNet121 emerged as a competitive alternative, though sensitivity was limited. The pretrained ResNet50 variant collapsed completely in the T4 task, with sensitivity of 0.000 despite AU-ROC of 0.880, reflecting extreme bias toward the majority class. These findings reinforce that task-specific simplicity, and anatomically constrained inputs may yield superior generalisability over deeper architectures, but also underscore the persistent challenge of detecting rare, advanced cases.

Beyond performance metrics, our Grad-CAM++ explainability experiments for the T4 classification task revealed that model attention was often anatomically plausible, with activations concentrated around peri-cartilage and paraglottic regions. However, CAM-mask overlap remained modest (Dice <0.1, enrichment ratios approximately 1.3–1.6), and mask-guided perturbations produced near-zero or negative probability changes, indicating weak causal coherence. Perturbation curves confirmed that salient voxels influenced predictions, but absolute probability shifts were small. These findings should therefore be interpreted only as exploratory and hypothesis-generating and require further validation on a larger scale.

From a translational standpoint, this work highlights three major implications. First, anatomically guided attention and segmentation-aware training may help constrain model focus to clinically relevant structures, reducing dependence on non-specific contextual cues. Second, integrating radiomic features such as shape, texture, and local intensity heterogeneity, with deep feature embeddings could improve the detection of subtle invasion by combining handcrafted sensitivity with DL's spatial representation power. Third, addressing data scarcity through synthetic augmentation (e.g., 3D GANs, variational autoencoders, and diffusion models) may balance rare T4 examples, mitigate overfitting, and provide more robust learning of pathological variation. These strategies collectively represent a path toward clinically reliable, anatomy-informed AI for cartilage invasion detection.

While the current models demonstrate limited sensitivity for T4 disease, this should not be viewed as a failure of the benchmark itself. Rather, it reflects the real-world difficulty of detecting subtle invasion patterns in CT imaging and validates the importance of having a standardised dataset to study this challenge. The benchmark's role is to expose such limitations transparently and catalyse innovation in model design and training strategies. This is a mere baseline benchmark and should not be considered as a near-deployable decision support tool. We acknowledge that substantial further methodological innovation and external validation is required before any T4 classifier could be considered for clinical decision support.

From a technical standpoint, this study highlights several methodological challenges that define clear and actionable research opportunities. A primary challenge is the severe class imbalance, with T4 tumours representing a small minority of cases. Despite focal-loss optimisation, sensitivity gains remained modest. Future work could explore advanced resampling strategies, GAN-based synthetic data generation, and semi-supervised learning with pseudo-labelling to better capture the morphological spectrum of invasion while preserving data realism. The observed differences between the default-threshold and optimised-threshold performance, together with the reported bootstrap confidence intervals for calibrated metrics, highlight the importance of reporting both discrimination and well-calibrated probabilities for safety-critical tasks such as T4 detection.

Although external validation using a leave-one-dataset-out (LODO) approach is a common generalisability strategy, we deliberately opted for a stratified multi-source validation across all six TCIA datasets. Each training and test subset was balanced to include samples from all sources and tumour T-stages, mitigating domain bias while maintaining statistical power. Given the substantial heterogeneity and profound T4 class imbalance across the six datasets, LODO evaluation would have yielded non-representative and statistically unstable results. Thus, the adopted stratified strategy provides a more realistic and statistically robust estimate of model generalisation across heterogeneous acquisition settings. Nevertheless, this design does not constitute true external validation, because both training and test sets were drawn exclusively from TCIA collections, which share acquisition practices and archival biases. As a result, real-world performance in prospective multi-centre deployments, especially in institutions with different scanners, protocols, and stage distributions, is expected to be lower than the estimates reported here, and future work should include genuine external test cohorts to more fully characterise generalisability.

Another limitation concerns generalisability across scanner types and contrast phases. The DeLong test revealed statistically significant variability between lightweight and deeper networks, suggesting that increased architectural depth does not necessarily improve robustness in heterogeneous datasets. Task-adapted simplicity, combined with cross-dataset calibration, may offer more reliable performance. These findings are consistent with the narrower confidence intervals and lower cross-validation standard deviations observed for the custom 3D CNN model, which indicate more stable behaviour across folds, whereas deeper residual networks exhibit larger variance and less reliable generalisation despite occasional high fold-wise scores. Because multiple pairwise AU-ROC comparisons were evaluated using the Delong's test without formal correction multiplicity, the very small p-values in the heatmaps are interpreted as qualitative support for the overall trends rather than definitive evidence of superiority.

Our explainability analyses suggested that model attention was often anatomically plausible and broadly aligned with radiological appearance but skewed toward non-invasive patterns, and these observations should be viewed as exploratory given the small number of analysed cases and modest CAM-mask overlap. This underscores the need for region-aware training pipelines, where segmentation masks or atlas priors guide feature learning toward biologically meaningful subregions (e.g., cartilage cortex, peri-cartilaginous structures). Future work could also integrate multi-modal inputs such as CT, MRI, PET, and clinical metadata to capture complementary signatures of invasion and tumour aggressiveness.

Future iterations of this benchmark should support hybrid modelling frameworks that fuse radiomic features with transformer or CNN-based deep representations. Such approaches have shown promise in other oncological domains by combining the interpretability of radiomics with the nonlinear representation capacity of deep learning, leading to improved calibration and clinical interpretability. While this benchmark focuses on classification tasks, future extensions may include segmentation-based models (e.g., 3D UNet) to support voxel-level staging and anatomical localisation, further enhancing clinical utility.

Beyond CT-based benchmarks such as LaryngealCT, parallel developments in oncology include theranostic radiopharmaceuticals and vitamin-functionalized nanocarriers that couple molecular imaging with targeted radionucleotide therapy, as well as DL methods for MRI-to-synthetic-CT generation that aim to reduce ionising radiation exposure during treatment planning[33,34]. In other disease domains, multimodal DL frameworks that integrate imaging with clinical or serological data have shown promise for non-invasive staging (for example in non-alcoholic fatty liver disease and generic throat cancer cohorts), although these studies typically rely on single-centre, non-public datasets without standardised benchmarking pipelines[35,36]. These directions are conceptually complementary to our work and suggest that future extensions of LaryngealCT could explore multimodal fusion and dose-sparing imaging strategies for laryngeal cancer staging, building on the curated volumes of interest and baseline models reported here.

By combining open access data, interpretable benchmarking, and clinically aligned performance metrics, LaryngealCT sets a new standard for reproducible and explainable AI development in laryngeal cancer staging. Crucially, its value as a benchmark is not solely in achieving optimal performance but in revealing clinical and technical bottlenecks, thus catalysing innovation in algorithmic design, training strategies, and clinical workflow integration.

# Methods

## Dataset Acquisition and Curation

We performed a targeted search of TCIA to identify public datasets containing CT scans of patients with head-and-neck cancer, including confirmed laryngeal cancer. Six datasets were downloaded after obtaining restricted-access licenses: (i) RADCURE[21], (ii) Head-Neck-PET-CT[22], (iii) HEAD-NECK-RADIOMICS-HN1[23], (iv) HNSCC[24], (v) Head-Neck-3DCT-RT[25], (vi) QIN-HEADNECK[26].

All imaging data and associated clinical metadata, including patient age, sex, tumour subsite, TNM staging, and treatment details, were retrieved using the National Biomedical Imaging Archive (NBIA) Data Retriever. This tool facilitates batch retrieval of imaging data and associated manifest files. Files were systematically organised into dataset-specific subfolders with linked metadata spreadsheets. After excluding poor-quality scans and post-laryngectomy cases, a final cohort of 1,029 cases was established from an initial pool of 1,043 studies. Table 2 provides the distribution of cases across datasets and T-stages. To ensure methodological rigour, we enforced strict patient-level separation between training and test sets, such that no patient contributed scans to more than one split. In addition, stratification was performed across the six TCIA sources to balance dataset representation and mitigate site-specific bias.

# Data Preprocessing

## Image Selection and Conversion

For each patient, the highest-quality diagnostic series was manually selected from the multiple files associated with each patient ID, prioritising diagnostic contrast-enhanced scans when available. The selected DICOM series were then converted to volumetric NRRD format using custom Python scripts to standardise input.

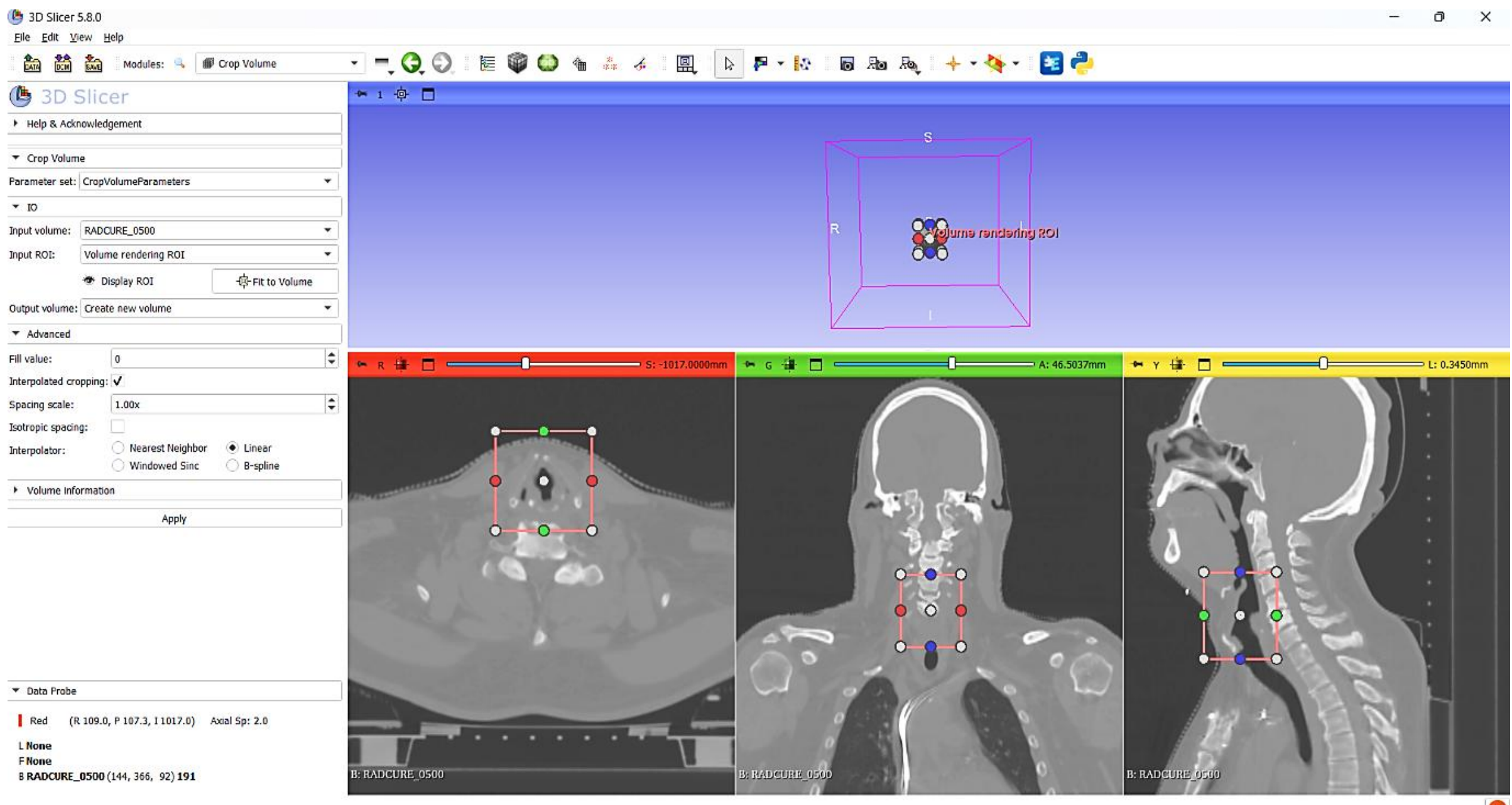

**Fig. 7 | Screenshot of 3D Slicer opened in the Crop Volume module.** Bounding boxes for the laryngeal crops are shown in red, with focus on laryngeal anatomy and the standardised reduction of the volume of interest (VOI). The left panel displays the default cropping parameters, including linear interpolation and 1 mm isotropic spacing.

## Cropping and Resampling of Laryngeal Volumes

Each image volume was loaded into 3D Slicer[37], an open-source tool used for medical image segmentation and visualised using the Volume Rendering module. The conventional viewing layout displays the 3D volume in the top row, with axial, coronal, and sagittal slices arranged from left to right in the bottom row. First, the region encompassing the larynx, defined from the superior edge of the epiglottis to the inferior boundary of the cricoid cartilage, was identified on the sagittal view. Next, the thyroid notch was located on the axial slice, and the boundaries were adjusted to ensure inclusion of the entirety of the laryngeal anatomy. Cropping was performed using the Crop Volume module in 3D Slicer, employing default settings including 1 mm isotropic voxels and linear interpolation, to maintain anatomical precision and uniformity across samples. Fig. 7 illustrates the Crop Volume user interface and bounding box selection in the axial, coronal, and sagittal planes, while Fig. 8 displays representative views of a cropped larynx volume.

Outputs were saved as NRRD files with the suffix Cropped_Volume.nrrd, reducing original image sizes by approximately 99.8%, thereby enabling efficient model training focused on the region of interest.

## Parameter Search for Reproducible Cropping

During the initial manual cropping of laryngeal CT volumes using 3D Slicer, sub-volumes corresponding to the laryngeal region were visually selected based on anatomical landmarks. To enable reproducibility and automated processing across the full dataset, it was necessary to retrospectively determine a standardised bounding box for each cropped region. We developed a parameter search strategy that systematically evaluated multiple sub-volume candidates within the original CT volume. Each candidate crop was compared to the manually extracted volume using a similarity metric, mean-squared error (MSE), to identify the best-matching region. The optimal bounding box for each scan was selected, and its coordinates were logged as origin and size parameters in x, y, and z dimensions. The coordinates of this optimal match were then stored as the bounding box for that case (bbox3d.txt). The outputs of this parameter search experiment were the cropping parameters and a reference image to compare the manually cropped and the automatically cropped first slice of the image. Metadata linking clinical details with imaging were compiled in LaryngealCT_metadata.xlsx. A Python utility (dataprep.py) was developed to crop raw TCIA data locally, ensuring reproducibility without redistributing derived volumes.

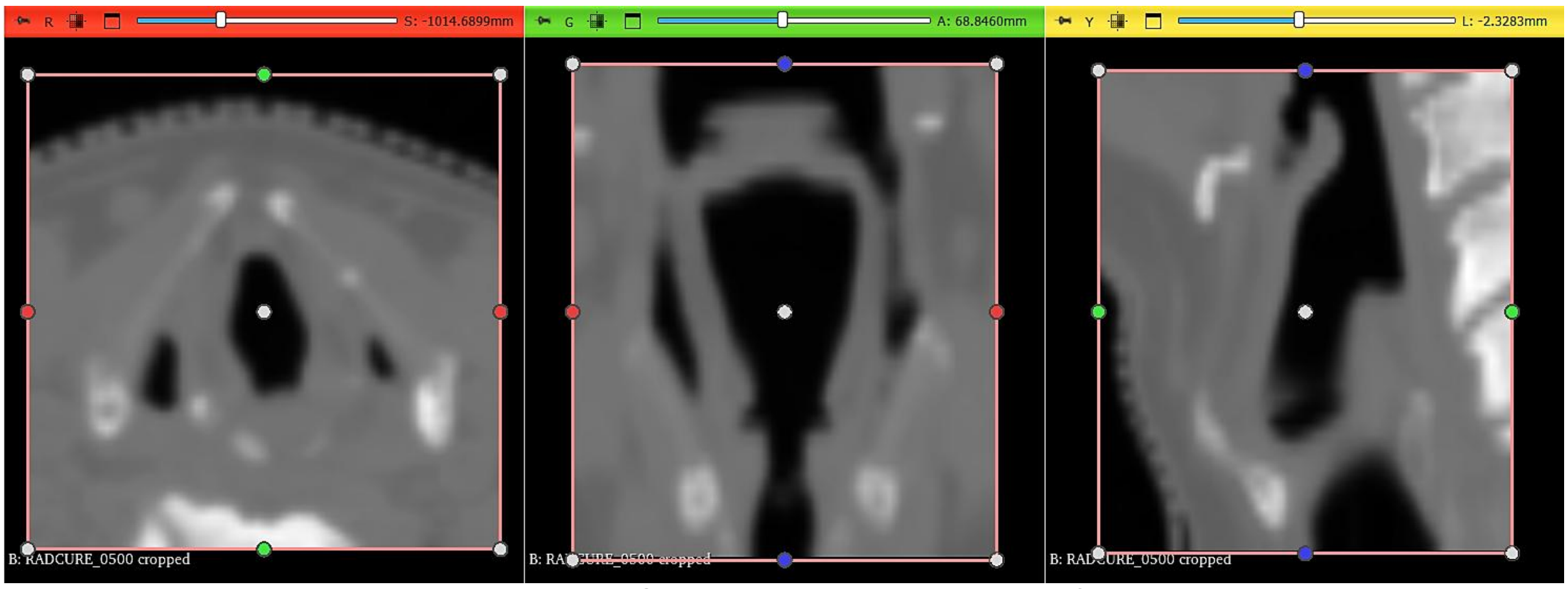

**Fig. 8 | Representative cropped laryngeal CT volume shown in axial (left), coronal (middle), and sagittal (right) planes.** The red bounding box denotes the standardised crop obtained through the parameter-search procedure, ensuring consistent anatomical coverage of the larynx across cases.

The parameter search produced case-specific bounding box coordinates (bbox3d.txt), enabling reproducible laryngeal cropping. To verify the fidelity of this approach, a dedicated quantitative validation step was performed. This method allowed us to translate initial expert-guided selections into a reproducible and automatable cropping pipeline, preserving the anatomical precision of manual annotations while ensuring that subsequent DL models could rely on consistent region-of-interest extraction across the entire dataset.

## Metadata Management, Dataset Organization and Reproducibility

Due to TCIA's license restrictions, the derived cropped volumes could not be redistributed directly. Instead, we developed an open-source reproducibility framework enabling researchers to reconstruct identical cropped datasets using their local copies of raw TCIA data**.** This process, illustrated in Fig. 9, includes:

**TCIA data download:** Six different datasets containing laryngeal cancer obtained with a restricted-access license.

**Create the data folder:** The folder structure is detailed as follows. The "data" folder is the main folder. A subfolder named "annotations" is used to store the "bbox3d.txt" file containing the cropping parameters, "imglist.txt" file containing the list of laryngeal cancer images from TCIA, and an Excel sheet "LaryngealCT_metadata.xlsx" containing the clinical details. The subfolder "tcia" contains the CT images from the six TCIA datasets (in separate folders) in NRRD format. The "cropped_nrrd" is the folder to save the cropped volumes.

**Crop laryngeal volume:** A Python script named "dataprep.py" is used to crop the laryngeal 3D volumes from the original TCIA images. All functions and documentation are available in the accompanying GitHub repository (https://github.com/funzi-son/LaryngealCT ).

**Data quality control:** Each case was systematically reviewed for completeness of metadata and adequate image quality. Scans with substantial artifacts, missing clinical data, or ambiguous anatomical landmarks were excluded from the final dataset.

**Metadata mapping:** Clinical metadata, including demographic and staging information, were mapped to each cropped CT volume and indexed in the “LaryngealCT_metadata.xlsx” spreadsheet located in the “annotations” subfolder, facilitating harmonized downstream analyses.

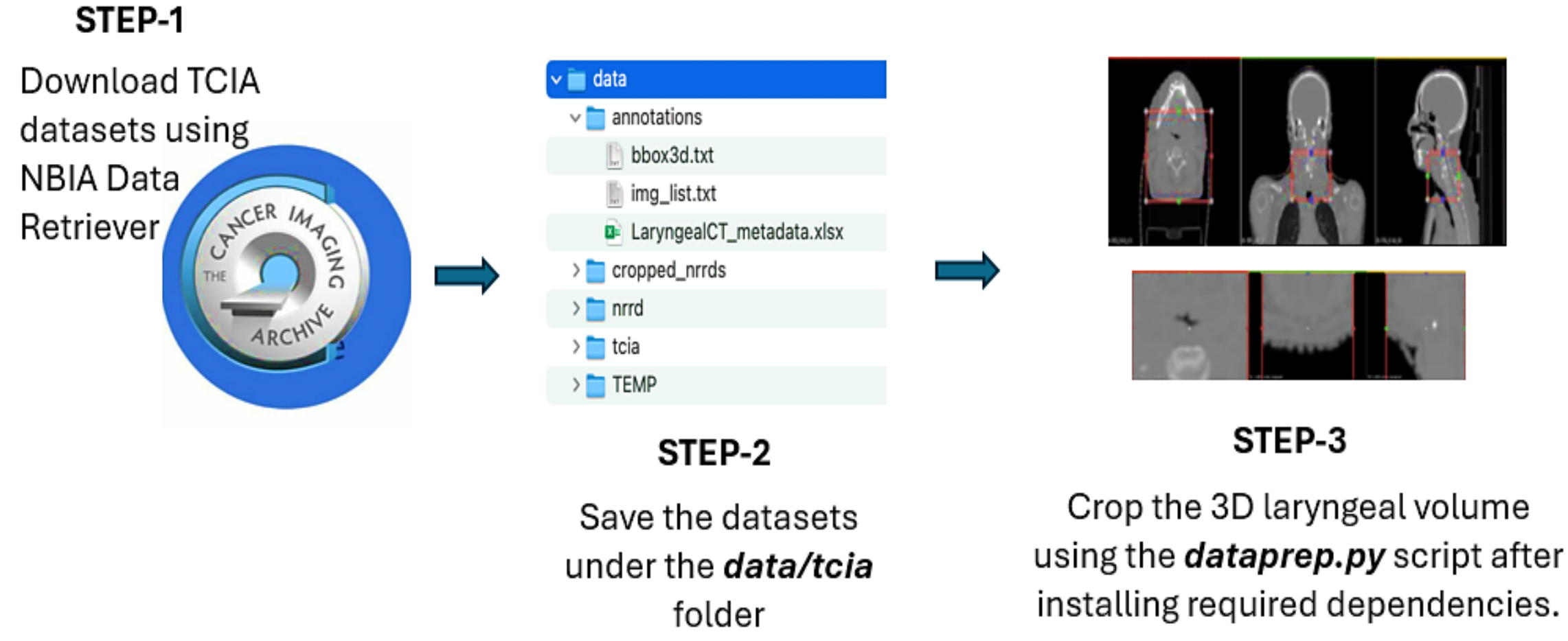

**Fig. 9 | Workflow illustrating example usage.** This illustrates the three steps for reproducing the LaryngealCT dataset from TCIA collections.

## Validation of Cropping

To evaluate both the clinical and computational validity of the preprocessing pipeline, we performed complementary qualitative and quantitative assessments.

**Qualitative validation:** A subset of 100 randomly selected cropped volumes was independently reviewed by two senior clinicians, including a board-certified head-and-neck surgeon and a radiologist. Each volume was scored on four criteria: (i) anatomical coverage, (ii) image quality, (iii) segmentation feasibility of the thyroid cartilage, and (iv) classification feasibility for T-staging. Scores were assigned on a 5-point Likert scale (1 = Poor to 5 = Excellent). Inter-rater reliability was quantified using weighted Cohen’s Kappa and raw agreement percentage with 95% confidence intervals.

**Quantitative validation:** To confirm the fidelity of automated crops derived from the parameter search, we compared them against all 1,029 expert-cropped reference volumes. Following resampling to 1 mm isotropic resolution, paired volumes were evaluated using Dice coefficient, Intersection-over-Union (IoU), mean squared error (MSE) and mean absolute error (MAE). These metrics confirmed that the automated cropping parameters reproduced expert crops with high accuracy across the full dataset.

## DL Benchmarking

### Model Selection and Rationale

Six 3D DL architectures were evaluated to benchmark the LaryngealCT dataset on two clinically significant classification tasks: (i) early (Tis–T1–T2) versus advanced (T3–T4) staging, and (ii) T4 versus non-T4 disease:

**Custom 5-layer 3D CNN:** A lightweight volumetric architecture designed specifically for this study to provide an interpretable and computationally efficient baseline. The model comprises five sequential convolutional blocks, each including a 3D convolution (kernel size 3×3×3), instance normalisation, ReLU activation, and MaxPooling for progressive downsampling. The final block applies an adaptive global average pooling layer, followed by a fully connected classifier with dropout regularisation (dropout rate=0.3) and a SoftMax two-class prediction. With approximately 1.5 million trainable parameters, this design balances efficiency with adequate capacity to capture hierarchical anatomical features of the larynx. The custom 3D CNN architecture is illustrated in Fig. 10.

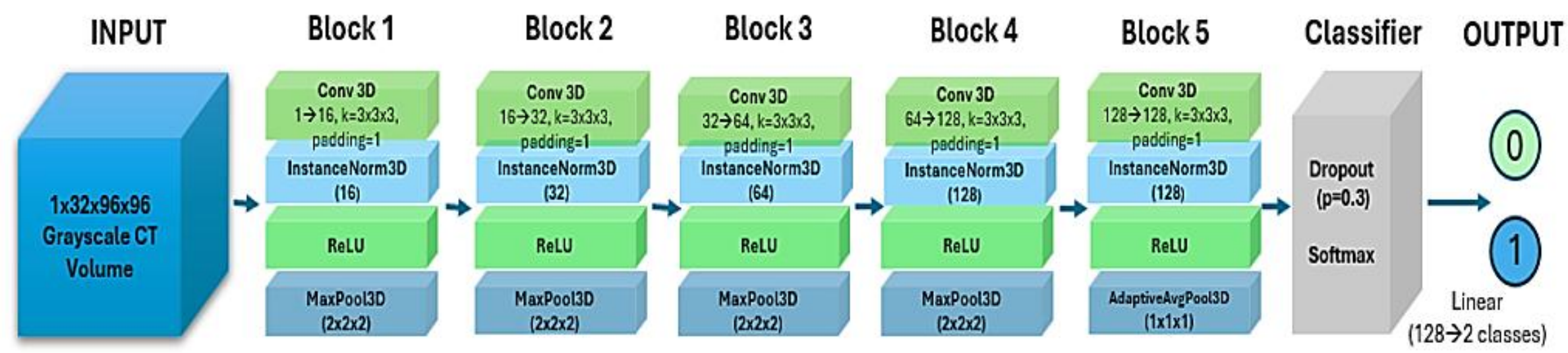

**Fig. 10 | Architecture of the custom 3D convolutional neural network (CNN) used for binary classification.** The model comprises five sequential 3D convolutional blocks (kernel size = 3, padding = 1), each followed by Instance normalisation, ReLU activation, and max pooling. The final block includes adaptive average pooling, and the classifier section applies dropout (p = 0.3) and SoftMax for two-class prediction.

**ResNet18:** A shallow residual network employing skip connections to mitigate vanishing gradients, prioritised for generalisation and reduced risk of overfitting on a mid-sized dataset.

**ResNet50:** A deeper residual variant with bottleneck blocks, balancing representational power and computational cost.

**ResNet101:** A high-capacity residual model designed to evaluate whether increased depth yields performance gains or overfitting.

**DenseNet121:** A densely connected architecture promoting feature reuse and efficient gradient propagation, potentially enhancing performance in limited-data settings.

**MedicalNet-pretrained ResNet50:** A ResNet50 variant initialized with MedicalNet weights pretrained on large-scale medical image datasets, enabling evaluation of transfer learning benefits for laryngeal cancer staging.

These models were chosen to represent a range of network complexities and design philosophies, facilitating a comprehensive evaluation of model capacity versus generalisability on a mid-sized, anatomically focused dataset. These architectures have been previously validated in the context of laryngeal cancer[13–15,17,38], underpinning their suitability for this task.

Additionally, transfer learning with MedicalNet-pretrained ResNet50 was explored to test the value of pretraining on large multi-organ medical datasets. Full fine-tuning with discriminative learning rates was explored, enabling assessment of whether domain-specific pretraining enhances laryngeal cancer staging. This diverse selection enables investigation of architectural impacts on predictive accuracy, convergence stability, and computational cost, providing insights critical for deploying clinically viable models in resource-constrained environments.

## Preprocessing and Data Augmentation

All CT volumes underwent standardised preprocessing, including fixed HU windowing, intensity normalisation, and spatial resampling, to reduce inter-scan variability and ensure compatibility with the deep learning architectures. First, voxel intensities were clipped to the range −300 to 300 Hounsfield Units (HU) to suppress irrelevant soft-tissue and high-density bone signals, thereby emphasising the diagnostic window most relevant for laryngeal structures. Each volume was then normalised using a z-score transformation to standardise intensity distributions across patients and scanners.

Cropped laryngeal volumes were resampled to a fixed spatial resolution of 32 × 96 × 96 voxels. This resolution was empirically chosen to balance two competing requirements: (i) sufficient coverage of the laryngeal region, spanning from the epiglottis to the cricoid cartilage, while preserving critical anatomical details such as thyroid cartilage margins; and (ii) computational efficiency, ensuring models could be trained within the memory limits of contemporary GPUs without sacrificing input fidelity.

To enhance model generalisation and reduce the risk of overfitting, on-the-fly data augmentation was applied to the training set using the TorchIO library. The following transformations were implemented with specified probabilities:

**Random left–right flipping** (p = 0.5), exploiting natural anatomical symmetry of the larynx.

**Random affine transformations** (scales 0.95–1.05, rotations ±8°, translations ≤ 2 voxels; p = 0.3), simulating positional and orientation variability during imaging.

**Random gamma correction** (log range −0.2 to 0.2; p = 0.2), mimicking variations in image contrast and brightness

**Gaussian noise injection** (σ ≤ 0.02; p = 0.15, increasing robustness against scanner-related artifacts.

These augmentations were applied dynamically during training, generating a diverse set of plausible input variations without inflating dataset size. This strategy improved model robustness across heterogeneous imaging conditions and enhanced the reproducibility of downstream classification performance.

## Training Protocol

DL models were trained end-to-end from randomly initialised weights, with additional transfer learning experiments conducted for ResNet50 using publicly available MedicalNet weights. Input volumes were preprocessed as described above and augmented during training. Training was performed using the AdamW optimizer (initial learning rate = $1 \times 10^{-4}$), with a Reduce-on-Plateau scheduler (factor = 0.3, patience = 3 epochs, minimum learning rate = $1 \times 10^{-6}$), monitoring the validation macro-F1 score.

To mitigate class imbalance, we employed focal loss (Equation 1) with task-specific tunable hyperparameters alpha and gamma, which downweights easy examples while emphasising misclassified cases.

$$\text{FocalLoss}(p_t) = -\alpha_t (1 - p_t)^{\gamma} log(p_t) \qquad (1)$$

where $p_t$ is the predicted probability of the true class, $\alpha_t$ adjusts class balance, and γ controls the focus on difficult samples. We chose α=0.75 and γ =2.0 for early versus advanced classification (Tis/T1/T2 vs. T3/T4), whereas for the T4 vs. non-T4 classification task, we opted for an increased alpha (α=8.0) with γ =2.0 following a grid search, reflecting the severe class imbalance.

Models were trained with an effective batch size of 8, achieved via gradient accumulation (2 samples × 4 steps), and optimised using mixed-precision training to reduce memory overhead.

Up to 500 epochs were permitted, with early stopping triggered if validation macro-F1 failed to improve for 20 epochs.

Data were split into 80% training (n = 823) and 20% independent testing (n = 206), stratified by stage. Within the training set, 5-fold stratified cross-validation was performed, under a fixed training setup to tune hyperparameters and assess stability. For the independent test cohort, predictions were generated by averaging probabilities across all five cross-validated models (ensemble), followed by a calibration and threshold optimisation derived from the validation folds. This ensemble protocol ensures that the test evaluation is independent of fold selection and avoids bias toward any single fold.

During training, the best model weights were saved per fold at the epoch yielding the highest validation macro-F1. For each validation fold, the predicted probabilities were calibrated using Youden's J statistic to identify the optimal decision threshold and temperature scaling to improve probability calibration. Final test performance was reported at three thresholds: (i) default (0.5), (ii) optimised (Youden's J), and (iii) calibrated (temperature scaling + optimised threshold).

In addition to training from scratch, we performed a transfer learning experiment with ResNet50 initialised from MedicalNet-pretrained weights. Full fine-tuning was applied, updating the entire network with discriminative learning rates (base layers at $3 \times 10^{-5}$, head at $1 \times 10^{-3}$). This experiment allowed us to assess the benefit of domain-specific pretraining for laryngeal cancer staging.

All experiments were conducted on a personal workstation equipped with an NVIDIA RTX 4060 GPU (8 GB VRAM), Intel Core i7 processor, and 24 GB RAM, running Windows 11 with CUDA 12.1, PyTorch 2.1, and MONAI 1.3. Mixed-precision training was enabled to reduce GPU memory overhead and accelerate computation. These specifications ensure that the pipeline is reproducible on widely available academic hardware rather than requiring access to large-scale high-performance cluster (HPC) resources. On the described workstation, the end-to-end inference time for the custom 3D CNN, including preprocessing and forward pass, was approximately 0.06-0.07 seconds per cropped laryngeal volume, while deeper residual architectures such as ResNet50/101 required roughly 0.08-0.10 seconds per volume.

## Evaluation Metrics

Model performance was comprehensively assessed using a diverse set of metrics designed to capture overall accuracy, class-specific performance, probability calibration, and clinical interpretability. Accuracy was reported as a general measure of predictive correctness; however, it can be misleading in imbalanced datasets, as it is dominated by the majority class[39]. To address this, balanced accuracy was also included, which averages the recall for each class and provides a more reliable assessment of model performance across both majority and minority classes.

Class-specific metrics such as precision, recall, and F1-score were reported per class, as well as macro-averaged and weighted, to specifically quantify sensitivity to minority classes (notably T3-T4 and T4) and to assess balance across categories. The Matthews correlation coefficient (MCC) was also included as a robust single-value metric that remains reliable even under skewed class distributions.

Discrimination ability was evaluated using two complementary metrics: the area under the receiver operating characteristic curve (AU-ROC) and the area under the precision-recall curve (PR-AUC). AU-ROC is appropriate for balanced datasets, measuring the trade-off between sensitivity and specificity, whereas PR-AUC provides additional insight by emphasising performance on the minority positive class, which is critical when positive cases are rare.

To enhance the clinical relevance and balanced performance of our binary classification model on imbalanced data, we applied both temperature scaling calibration and decision threshold optimisation. Temperature scaling is a post-hoc calibration technique that adjusts SoftMax logits using a scalar temperature parameter T, optimised on the validation set to minimise negative log-likelihood. This preserves class rankings while refining confidence estimates.

Threshold optimisation was performed by maximising Youden's J statistic (Equation 2) on the validation set predictions to jointly optimise sensitivity and specificity:

$$\text{Youden's J} = \text{Sensitivity} + \text{Specificity} - 1 \qquad (2)$$

Calibration of probabilistic outputs was assessed using the Brier score, which quantifies the mean squared difference between predicted probabilities and actual outcomes, thereby penalising both over- and under-confident predictions. This is formally defined as (Equation 3):

$$Brier\ Score = \frac{1}{N}\sum_{i=1}^{N}(f_i - o_i)^2 \qquad (3)$$

where $f_i$ denotes the predicted probability for case *i*, and $o_i$ is the observed ground truth label, and N is the total number of cases. A lower Brier score indicates better calibration, penalising both over- and under-confident predictions, with a perfect score being 0.

Following calibration, which adjusts model-predicted probabilities to more accurate likelihoods via temperature scaling, we evaluated model performance under three regimes: the default threshold of 0.5, the optimised threshold (t_opt) selected by Youden's J, and calibrated thresholds combining temperature scaling and threshold optimisation. We report performance metrics for both raw (non-calibrated) and calibrated (optimised) predictions, with 95% confidence intervals computed by 1000-sample patient-level bootstrap resampling on the independent test set. Presenting optimised metrics provides a more faithful assessment of the model's real-world applicability, particularly improving sensitivity and precision for the minority class.

To assess discrimination capacity independent of any decision threshold, receiver operating characteristic (ROC) and precision-recall (PR) curves were generated from raw predicted probabilities. ROC overlays were constructed by averaging validation fold-level ROC curves (mean ± standard deviation) and comparing them with the independent test ROC for each model. This dual reporting allows clear separation of threshold-specific performance (clinical utility) from overall separability of classes (discrimination ability).

## Statistical Analysis

For each model and task, 95% confidence intervals on test-set metrics (accuracy, balanced accuracy, F1-macro, sensitivity, specificity, AU-ROC) were obtained by 1,000-sample patient-level bootstrap resampling using the full set of test predictions. Pairwise AU-ROC comparisons between models were performed using DeLong's test on the test predictions, and the resulting p-values are visualised as heatmaps to highlight relative performance differences. In addition, Wilcoxon signed-rank tests were performed across cross-validation folds to compare AU-ROC and F1-macro distributions between models, complementing DeLong comparisons by assessing fold-wise variability. Because many pairwise comparisons were evaluated, these tests are treated as exploratory and are not adjusted for multiple testing; consequently, small p-values are interpreted qualitatively as supporting overall trend that lightweight architectures outperform deeper networks in this setting rather than as definitive evidence of superiority for any single model. For cross-validation results, performance was summarised as mean ± standard deviation across the five folds, and higher variance, particularly in deeper residual networks, is interpreted as reduced training stability on this mid-sized, heterogeneous dataset with pronounced class imbalance.

### Explainability Analysis

To ensure model transparency and clinical interpretability for the more clinically relevant T4 classification task, three complementary analyses were performed: Grad-CAM++ visualisation, perturbation-based sensitivity testing, and anatomical alignment assessment. Grad-CAM++ heatmaps were derived from the final convolutional layers of both the custom 3D CNN and the ResNet101 ensembles to localise discriminative 3D regions influencing the T4 versus non-T4 decision.

To evaluate causal relevance, deletion and insertion experiments were conducted by progressively removing or re-introducing voxels ranked by Grad-CAM++ intensity and quantifying the resulting change in model confidence. A complementary perturbation-based sensitivity experiment was performed within manually segmented, expert-validated thyroid cartilage masks to assess whether model predictions explicitly depended on cartilage voxels. Finally, Grad-CAM++ activations were compared with these cartilage masks using Dice similarity, coverage ratio, and activation-enrichment metrics to quantify spatial correspondence. Six representative cases (three T4 and three non-T4) were selected for detailed interpretability analysis. These analyses provide exploratory insights into peri-cartilage attention patterns but overlap metrics and perturbation curves are interpreted cautiously given the small sample size and modest Dice values.

# Ethical Considerations

Our work adheres strictly to data privacy and ethical standards. The original CT images were collected from The Cancer Imaging Archive (TCIA) under controlled-access NIH policies, with access obtained through the Restricted Access License process as required by TCIA and NIH. All requests for data were approved by TCIA, and usage fully complied with TCIA licensing terms and data usage restrictions. All research was performed in accordance with relevant institutional and national guidelines and regulations. Experimental protocols for this study were reviewed by the Deakin University Human Research Ethics Committee (Project ID: 2025HE000934), which confirmed that the use of the TCIA datasets and associated research activities are exempt from human ethics review. The requirement for informed consent was waived by the Deakin University Human Research Ethics Committee, as the research involved retrospective analysis of anonymized, publicly available data.

No attempt was made to re-identify patients from the datasets. Adhering to the TCIA Restricted Access licensing contract, we are not sharing any raw or cropped images through any supplementary material or our GitHub page. To harmonise open scientific progress with these constraints, we provide an open-source pipeline and comprehensive cropping parameters, allowing authorised researchers to reconstruct consistent, anatomically accurate VOIs from raw TCIA data. This approach promotes transparency, ethical compliance, and research reproducibility without compromising regulatory or licensing obligations.

# Data availability

The curated dataset is built upon six publicly available CT collections hosted by The Cancer Imaging Archive (TCIA): RADCURE [ DOI: 10.7937/J47W-NM11 ], Head-Neck-PET-CT [ DOI: 10.7937/K9/TCIA.2017.8oje5q00 ], HEAD-NECK-RADIOMICS-HN1 [ DOI: 10.7937/tcia.2019.8kap372n ], HNSCC [ DOI: 10.7937/k9/tcia.2020.a8sh-7363 ], HNSCC-3DCT-RT [ DOI: 10.7937/K9/TCIA.2018.13upr2xf ], and QIN-HEADNECK [ DOI: 10.7937/K9/TCIA.2015.K0F5CGLI ]. Access to these datasets is governed by the NIH controlled data access policy [https://www.cancerimagingarchive.net/nih-controlled-data-access-policy/] and requires registration, approval, and acceptance of TCIA's usage terms. Direct redistribution of raw CT scans is not permitted.

All derived components (cropping parameter files, clinical metadata, and train-test splits) are openly available via GitHub: https://github.com/funzi-son/LaryngealCT/tree/main/data.

# Code availability

The full preprocessing toolkit, automated cropping scripts, benchmarking pipelines, and trained model checkpoints are provided at https://github.com/funzi-son/LaryngealCT. This repository includes detailed instructions to reproduce all experiments reported in this study.

# Acknowledgments

We thank TCIA for granting access to the restricted datasets used in this study. We also acknowledge the computational support provided by the institutional GPU cluster at the School of Information Technology, Deakin University.

# Funding

This study did not receive any funding.

# Author Contributions

N.R. Conceptualisation, Dataset acquisition and curation, Methodology, Conducting the experiments, Analysis of results, Drafting of original manuscript, Preparation of figures, Creating GitHub repository. S.N.T. Conceptualisation, Technical supervision for coding, Evaluation of experiments, analysis and results, Reviewing and editing of manuscript, Creating GitHub repository. A.S. Conceptualisation, Overall supervision and strategic guidance, Evaluation of experiments, analysis and results, Reviewing and editing of manuscript. K.D. Conceptualisation, Expert validation of cropped VOIs, Overall supervision, Evaluation of experiments, analysis and results, Reviewing and Editing of manuscript. P.K. Conceptualisation, Expert validation of cropped VOIs, Supervision, Evaluation of analysis and results, Reviewing and editing of manuscript. Y.X. Conceptualisation, Technical supervision, and validation, Reviewing and editing of manuscript. D.R. Conceptualisation, Technical supervision and analysis of results, Reviewing and editing of manuscript. All authors reviewed and approved the final manuscript.

# Competing interests

All authors declare no financial or non-financial competing interests.